%% file: cvpr25_arxiv.tex
\documentclass[10pt,twocolumn,letterpaper]{article}

\usepackage{cvpr}              


\definecolor{cvprblue}{rgb}{0.21,0.49,0.74}
\usepackage[pagebackref,breaklinks,colorlinks,allcolors=cvprblue]{hyperref}
\usepackage{comment}
\usepackage{amsmath}
\usepackage{pifont}
\usepackage{amssymb}

\def\paperID{17583} 
\def\confName{CVPR}
\def\confYear{2025}

\title{ReSpec: Relevance and Specificity Grounded Online Filtering for \\ 
Learning on Video-Text Data Streams}

\author{
Chris Dongjoo Kim$^{1,2}$\thanks{These authors contributed equally to this work.},\quad Jihwan Moon$^{1}$\footnotemark[1],\quad Sangwoo Moon$^{1}$,  \quad
  Heeseung Yun$^{1}$,\quad Sihaeng Lee$^{2}$,\\ Aniruddha Kembhavi$^{3}$,\quad Soonyoung Lee$^{2}$,\quad Gunhee Kim$^{1}$,\quad Sangho Lee$^{3}$\thanks{Equal corresponding authors.},\quad Christopher Clark$^{3}$\footnotemark[2]  \\
  $^{1}$Seoul National University, $^{2}$LG AI Research, $^{3}$Allen Institute for AI\\
  {\small \texttt{\{cdjkim, jihwan.moon, sangwoo.moon, heeseung.yun\}@vision.snu.ac.kr}} \quad   {\small \texttt{gunhee@snu.ac.kr}} \\
  {\small \texttt{\{soonyoung.lee, sihaeng.lee\}@lgresearch.ai}}\quad {\small \texttt{\{chrisc, sanghol, anik\}@allenai.org}}
}

\begin{document}
\maketitle
\input{sec/0_abstract}

\input{sec/1_intro}
\input{sec/2_related_works}
\input{sec/approach}

\input{sec/3_experiments}

\input{sec/4_conclusion}

\input{sec/5_acknowledgments}

{
    \small
    \bibliographystyle{ieeenat_fullname}
    \bibliography{cvpr25_arxiv}
}

\input{sec/6_supplementary}

\end{document}

%% file: sec/0_abstract.tex
\begin{abstract}

The rapid growth of video-text data presents challenges in storage and computation during training. 
Online learning, which processes streaming data in real-time, offers a promising solution to these issues while also allowing swift adaptations in scenarios demanding real-time responsiveness.
One strategy to enhance the efficiency and effectiveness of learning involves identifying and prioritizing data that enhances performance on target downstream tasks. 
We propose \textbf{Re}levance and \textbf{Spec}ificity-based online filtering framework (\textbf{ReSpec}) that selects data based on four criteria: 
(i) modality alignment for clean data, (ii) task relevance for target focused data, 
(iii) specificity for informative and detailed data, and (iv) efficiency for low-latency processing. 
Relevance is determined by the probabilistic alignment of incoming data with downstream tasks, while specificity employs the distance to a root embedding representing the least specific data as an efficient proxy for informativeness.
By establishing reference points from target task data, ReSpec filters incoming data in real-time, eliminating the need for extensive storage and compute.
Evaluating on large-scale datasets WebVid2M and VideoCC3M, ReSpec attains state-of-the-art performance on five zero-shot video retrieval tasks, using as little as 5\% of the data while incurring minimal compute.
The source code is available at \href{https://github.com/cdjkim/ReSpec}{https://github.com/cdjkim/ReSpec}.
\end{abstract}

%% file: sec/1_intro.tex
\section{Introduction}
\label{sec:intro}

%

The world generates an immense volume of multimodal data, with video-text data in particular presenting unique challenges for the storage and training of machine learning agents. 
Video data requires substantial computational resources for both storage and training, due to its dynamic nature and high volume. 
This complexity necessitates an online framework capable of learning and processing data on-the-fly, without extensive storage requirements.

In this work, we argue that for a learning agent to perform effectively in an online environment, it must filter and prioritize data that enhances its downstream task objectives. 
Efficient data selection is crucial to streamline the online learning process, improve model accuracy, and reduce unnecessary computational loads.

\begin{figure}[t]
    \begin{center}
    \includegraphics[width=\columnwidth]{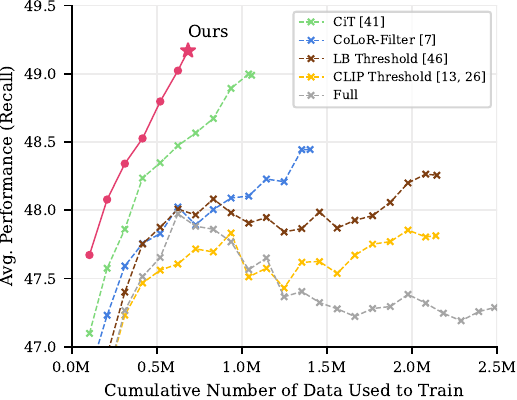}
    \caption{\textbf{Online training with online filtering.}
    Comparison of average performance across various methods for online filtering, shown as a function of the cumulative data samples used for training. 
    Our proposed method consistently outperforms baseline approaches \textbf{at any number of training iterations} on WebVid2M~\cite{bain21webvid}, achieving the \textit{highest performance} with \textit{minimal data} requirements.
    }
    \label{fig:teaser}
    \end{center}
    \vskip -0.2in
\end{figure}

An effective online filtering method should consider four essential criteria. 
\textit{(i)} Modality-Aligned Data Selection (Noise Filtering): It must ensure that selected data is correctly aligned across modalities (\ie, video paired accurately with text).
\textit{(ii)} Task-Relevant Data Prioritization: The method should prioritize data that directly contributes to the performance of specific downstream tasks, focusing the learning on the most relevant information. 
\textit{(iii)} Informative Data Collection: The system should identify and select data that provides high-value information. 
\textit{(iv)} Efficient Filtering: The method should allow for seamless, low-latency filtering of incoming data streams, minimizing computational overhead. 

Although substantial research has focused on multimodal dataset filtering in offline settings~\cite{gadre2023datacomp,wang2024normsim,maini2024tmars,kim2024hype}, there is a clear gap in methods designed for real-time online filtering, as illustrated in Fig.~\ref{fig:online_offline}. 
This gap is especially critical in applications demanding rapid, precise adaptations, such as Autonomous Driving: Vehicles must adapt to continually changing road conditions and respond to unexpected obstacles in real-time~\cite{muhammad2020deep,yang2023efficient} or 
Real-Time Surveillance: Continuous video streams need instant analysis to detect security threats or anomalous behaviors~\cite{torres2019online,kwon2022toward}. 
To the best of our knowledge, this work is the first to address the unique challenges of video-text data in an online filtering setting, where computational constraints and the dynamic nature of multimodal streams demand a highly efficient, task-driven approach.
Beyond meeting the need for fast adaptations, online approaches have a key advantage over offline methods when the goal is to curate data for training. 
They significantly reduce the storage and computational costs of handling video data, making them indispensable in resource-constrained settings with limited compute and storage capabilities.

Our \textbf{Re}levance and \textbf{Spec}ificity-based online filtering framework (\textbf{ReSpec}) begins by pre-processing target task data to obtain the representations, establishing reference points for evaluating relevance and specificity. 
Then, for each incoming data point, we first check the alignment between video and text content using a similarity score, ensuring cross-modal consistency and cleanness.
Next, we determine the data's relevance to downstream tasks via density estimation, evaluating its probabilistic relation to the target task distributions. 
This approach ensures that the filtered data aligns closely with the distributional properties required for enhancing task performance, enabling the framework to prioritize data that directly improves downstream objectives.
Finally, we assess the specificity of each sample by measuring its distance to a root embedding (\ie, text embedding of empty text `` '', representing the least specific embedding). 
This provides an efficient proxy for evaluating informativeness in an online setting, allowing the framework to filter out uninformative or generic samples.

The central question of our work is how effectively online filtered streaming data can enhance performance on five video retrieval zero-shot downstream tasks. 
As shown in Fig.~\ref{fig:teaser}, ReSpec consistently outperforms all baseline approaches across any training iterations while significantly reducing data usage to approximately 27\% for WebVid2M\cite{bain21webvid} and 5\% for VideoCC3M~\cite{nagrani2022videocc3m} (Fig.~\ref{fig:top1_perf}).
In addition to achieving the best average downstream task performance, ReSpec demonstrates robustness to hyperparameter settings (Fig.~\ref{fig:lb_threshold_analysis}, Tab.~\ref{tab:relevance_modality_ablation}) and stands out as the most computationally efficient method compared to all baselines (Tab.~\ref{tab:computation_cost}).

%% file: sec/2_related_works.tex
\section{Related Works}
\label{sec:related_works}
\subsection{Data Filtering on Noisy Dataset}
\label{sec:data_filtering}

Data filtering aims to improve model performance by filtering out data while enhancing storage and training efficiency. 
DataComp~\cite{gadre2023datacomp} is a prominent image-text dataset benchmark in this area.
However, most existing studies are either image-specific or offline-based, which introduces limitations when applying them to online video-text filtering.
T-MARS~\cite{maini2024tmars} replaces CLIP cosine similarity with a text-masked alignment score to reduce biases caused by text in images.
However, this approach is unsuitable for online video-text datasets due to the high visual masking costs.
CLIPLoss + NormSim~\cite{wang2024normsim} addresses the shortcomings of traditional CLIP cosine similarity-based data ranking by using a negCLIPLoss score that imitates the training loss. 
It also uses normalized similarity with downstream data as an additional metric for data ranking.
HYPE~\cite{kim2024hype} utilizes hyperbolic image-text model MERU~\cite{desai2023hyperbolic} along with CLIP to measure image and text specificity.
SIEVE~\cite{anas2024sieve} measures the alignment between the original text and text generated from the corresponding image to assess image-text alignment for filtering, but it requires a captioner, making it impractical for online use.

These approaches are primarily designed for offline settings, where the entire dataset is stored.
As such, they tend to entail high storage and computation costs, 
making them unsuitable for addressing the challenge of filtering video-text data from an infinite data stream in an online manner.

\subsection{Online Learning and Filtering}
\label{ec:online_learning}

Traditional offline learning involves training a model using the entire dataset all at once in an offline setting, which leads to inefficiencies in computational and storage cost.
On the other hand, online learning processes data sequentially, updating the model incrementally as new data arrives. 
This approach overcomes the limitations of offline learning by being more efficient and scalable, making it well-suited for large-scale, real-time applications where data is both extensive and rapidly incoming.

Online learning has largely been explored in the context of continual learning, which studies the problem of learning sequential tasks while alleviating catastrophic forgetting of past tasks~\cite{mai2022online,aljundi2019gradient,wang2022ocl-nds,cai2021cvt}.
Online learning has also been used to improve the training efficiency by specializing a model for specific target tasks~\cite{mullapudi2019omd,boldo2023query,farhadi2020enabling}.

Recently, downstream task-aware online filtering approaches have been introduced for efficient online training to boost performance on target downstream tasks.
CiT~\cite{xu2023cit} has shown that dynamically curating the training data based on its distance to target downstream tasks improves training efficiency of visual-linguistic multimodal models.
CoLoR-Filter~\cite{brandfonbrener2024color-filter} collects training data from an online text data stream by exploiting the log probability differences between a pretrained prior model and a downstream task fine-tuned model, achieving competitive downstream task performance while reducing the training data by up to 11$\times$.

While these works highlight the importance of downstream task-aware online filtering, they have shortcomings in computational efficiency, as CiT requires backpropagation using the filtered data and CoLoR-Filter requires two forward passes for each incoming data.
We propose a more efficient online filtering framework while incorporating the distributional aspect of the downstream task and the specificity aspect of the incoming data.

\subsection{Visual-Language Dataset Curation}
\label{sec:data_curation}
Generally, to improve the quality of noisy visual-language datasets collected from the web on a large scale, 
a pretrained visual-language foundation model is used to measure inter-modal similarity, 
applying appropriate thresholding.

For image-text datasets like LAION-5B~\cite{schuhmann2022laion-5b} and LAION-400M~\cite{schuhmann2021laion-400m}, 
web data was filtered using the cosine similarity of a pretrained CLIP~\cite{clip}, 
with thresholds of 0.28 and 0.3, respectively.
This technique is also applied to video-text datasets, 
as seen in Koala~\cite{tan2024koala}, where the video-text dataset (HowTo100M~\cite{miech2019howto100m}) 
was filtered based on a maximum frame-text CLIP cosine similarity of 0.26.
Recent studies (InternVid~\cite{wang2024internvid}, Panda-70M~\cite{chen2024panda}) 
have introduced text generation-based data curation methods using captioning models or large language models (LLMs).

While the curation approaches mentioned above can be applied in online filtering, they are limited in that they do not collect or generate data specifically needed for target downstream tasks but instead rely on the training data distribution of pretrained models.
This means that the collected or generated data is similar to the pretrained model's training distribution, which may not directly lead to improved performance in downstream tasks.

%% file: sec/approach.tex
\section{Method}
\label{sec:approach}

\begin{figure}[t]
    \begin{center}
    \includegraphics[width=\columnwidth]{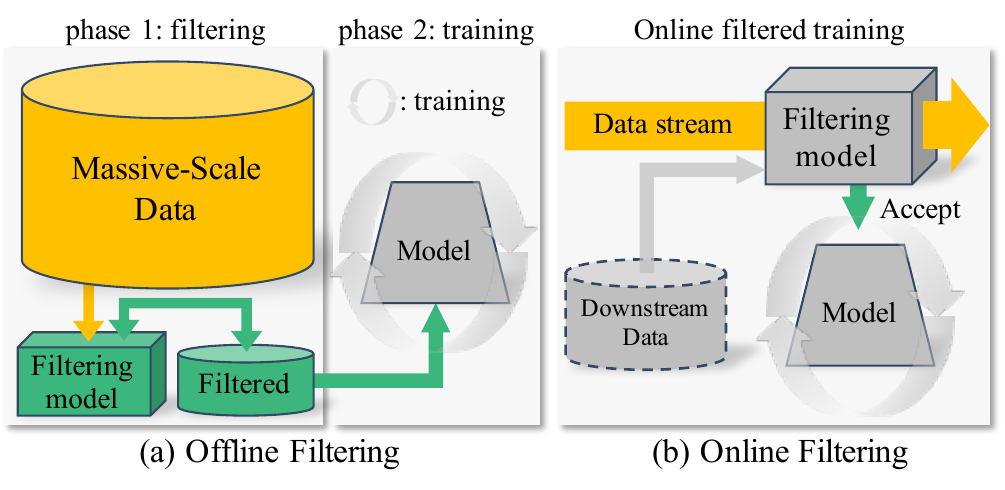}
    \vskip -0.1in
    \caption{\textbf{High-level comparison of offline and online filtered training.}
    (a) Offline filtering approaches initially process the stored massive-scale data (often by scoring and ranking the entire data) and retain the filtered subset. 
    The filtered data is then used during the training phase.
    (b) Online filtering performs real-time filtering, dynamically deciding whether to accept or discard samples. Accepted samples are immediately forwarded for online model training.
    Downstream task-aware online filtering, such as CiT~\cite{xu2023cit}, CoLoR-Filter~\cite{brandfonbrener2024color-filter}, and our ReSpec, utilizes downstream task data (e.g., embeddings) to guide the filtering process.
    }
    \label{fig:online_offline}
    \end{center}
    \vskip -0.3in
\end{figure}

As illustrated in Fig.~\ref{fig:online_offline}-(b), the online filtered multimodal learning problem addresses the need to process a continuous stream of data containing both video and text information. 
We define these paired data streams as $\{(v_t, s_t)\}_{t=1}^\infty$, where $v_t$ represents video data and $s_t$ is the associated text data at time $t$.
ReSpec aims to filter this data stream to online train a multimodal model capable of generalizing across multiple target downstream tasks, denoted as $\{\mathcal{D}_d\}_{d=1}^D$.
Each task dataset is represented by a set of paired features $\{(v_{d,i}, s_{d,i})\}$ where $v$ and $s$ are unit-norm video and text features extracted from a pre-trained visuo-text model.
For notation clarity, all instances of $v$ and $s$ hereafter represent features extracted from this pre-trained visuo-text model, and let $x$ denote features from either modality for cases that could apply to both $v$ or $s$.

The online filtering focuses on three key aspects.
\textit{Alignment} between the video and text modalities (ensuring data cleanness), \textit{relevance} to the target downstream tasks, and \textit{specificity} of the data, ensuring it is detailed enough for effective task learning.
These criteria guide the selection of data samples that contribute effectively to the model's performance on downstream tasks.

\begin{figure*}[h]
    \begin{center}
    \includegraphics[width=1.0\textwidth]{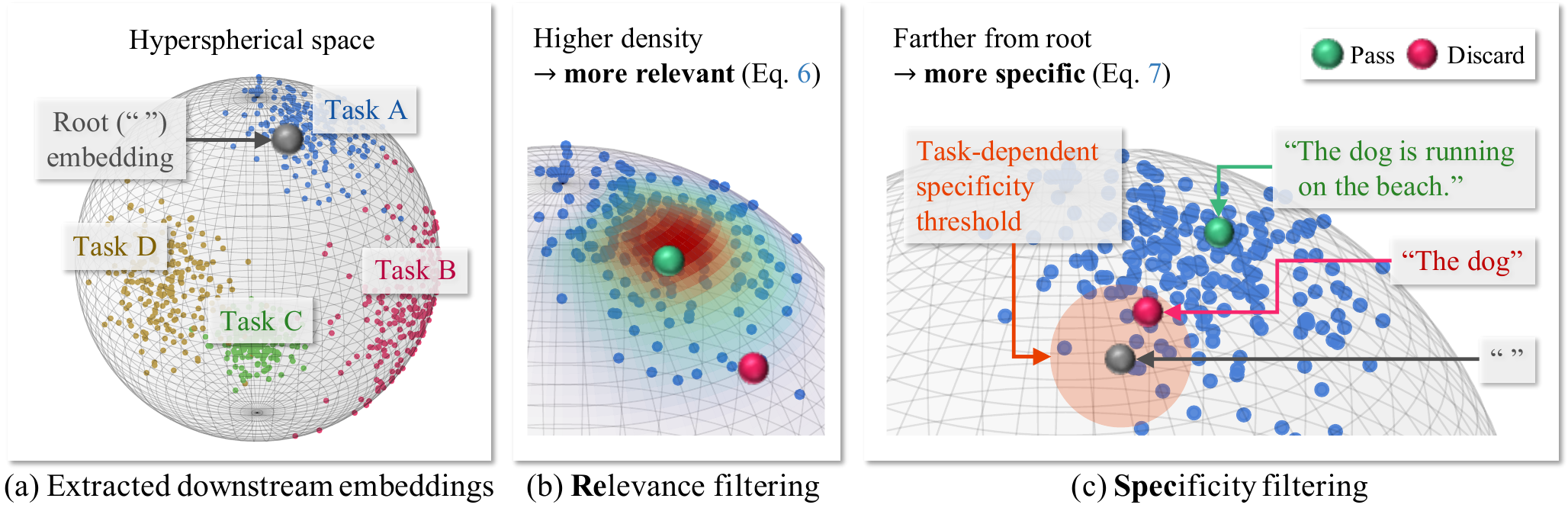}
    \vskip -0.05in
    \caption{\textbf{ReSpec: Relevance and Specificity based online filtering.}
    (a) We precompute downstream embeddings by utilizing the target downstream task dataset.
    (b) \textbf{Re}levance is determined by evaluating how closely a data point aligns with the target downstream embedding distribution using density estimation.
    (c) \textbf{Spec}ificity is measured by comparing the relative distances of the incoming embedding and the downstream embedding to the root embedding (empty text embedding, \ie, `` ").
    Video-text pairs that \textit{pass} all of the alignment, relevance, and specificity filters are \textit{accepted} for online model training.
    }
    \label{fig:filtering_framework}
    \end{center}
    \vskip -0.2in
\end{figure*}

As illustrated in Fig.~\ref{fig:filtering_framework}, this process begins with pre-processing to obtain representations of all target task data to establish a baseline for assessing relevance and specificity.
Then for each incoming data point, video and text information alignment is first assessed via similarity score, ensuring cohesive and clean data across modalities (\S~\ref{subsec:multimodal_alignment_filtering}).
Then, we determine the relevance to each target downstream task using density estimation of downstream features, allowing the system to prioritize data closely related to the tasks at hand (\S~\ref{subsec:relevance_filtering}).
Lastly, the specificity of the data is verified by comparing the distance from the root text embedding (\ie, embedding of empty text `` '') to the incoming sample and to the downstream task embeddings (\S~\ref{subsec:specificity_filtering}).
If the incoming data satisfies all of these three criteria on any downstream task, it is then used for the online training.
This structured filtering approach enables efficient and task-targeted training for multimodal models, enhancing performance on the target downstream tasks.

\subsection{Multimodal Alignment Filtering}
\label{subsec:multimodal_alignment_filtering}
Firstly, to assess multimodal alignment, we employ the commonly used technique of similarity score thresholding~\cite{schuhmann2021laion-400m,schuhmann2022laion-5b,tan2024koala,gadre2023datacomp}. 
The similarity score is calculated between the video and text modalities in each data sample, allowing us to filter for samples whose contents are well-aligned across both modalities. 
This ensures that only samples with strong cross-modal consistency are retained for further processing.

The incoming sample passes the multimodal alignment filter if
\begin{equation}
    \langle v_t, s_t \rangle > \tau,
\end{equation}
where $v_t$ and $s_t$ are the embeddings of the incoming sample's video and text components, respectively, $\langle \cdot , \cdot \rangle$ is the dot product, and  $\tau$  is a similarity threshold. 

\subsection{Relevance Filtering}
\label{subsec:relevance_filtering}

Our objective is to filter data that is relevant to one or more downstream tasks. 
Viewing the online data stream as the training distribution and the downstream tasks as the test distribution, it is both empirically and theoretically established that aligning the training distribution with the test distribution enables better generalization of the trained model on test data.
Theoretically, the generalization ability of a trained model is correlated with various divergence and distance metrics between the training and test distributions~\cite{wu2020information,jose2021information,zilly2019frechet}. 
Empirically, selectively aligning training data with the test distribution has been shown to enhance test performance while reducing computational costs~\cite{mahajan18,xu2023cit,brandfonbrener2024color-filter,xie23dsir}.
This also aligns with the No Free Lunch theorem~\cite{wolpert1996nft,wolpert1997nft}, which implies that no single representation is universally optimal across all data types and tasks. 
Tailoring representations to meet specific task requirements, therefore, is a promising strategy for improving model performance and efficiency.

To achieve this, we implement a method to represent each downstream task and evaluate the relevance of incoming data samples. 
This involves calculating the kernel density estimate for each incoming data sample in relation to each downstream task, using a von Mises-Fisher (vMF) kernel suited for unit-norm representations.
The vMF kernel is defined as:
\begin{equation}
    f_{\text{vMF}} (x_t ; \mu, \kappa) = C_z(\kappa) \exp (\kappa x_t^\top\mu),
\label{vmf_kernel}
\end{equation}
where $z$ is the dimension, $C_z(\kappa)=\frac{\kappa^{z/2-1}}{(2\pi)^{z/2}I_{z/2-1}(\kappa)}$ is the normalization constant, $\mu$ is the mean direction, and $\kappa$ is the concentration parameter.
Here, $I(\cdot)$ represents the modified Bessel Function of the first kind.

We use the Maximum Likelihood Estimation approximation of \cite{banerjee2005clustering} to obtain downstream task-specific $\hat{\kappa}$:
\begin{align}
    R=||\bar{x}_d&|| = \left | \left | \frac{1}{N_d} \sum_{n=1}^{N_d} x_{d,n} \right | \right | \\
    \hat{\kappa} &= \frac{R(z-R^2)}{1 - R^2},
\label{eq:kappa_estimation}
\end{align}
Then, the kernel density estimate at input embedding $x_t$ is:
\begin{equation}
    \hat{f}_{\text{vMF-KDE}}(x_t) = \frac{1}{N_d}\sum_{n=1}^{N_d} f_{\text{vMF}} (x_t ; x_{d, n}, \hat{\kappa}),
\label{vmf_kde}
\end{equation}
Here, $N_d$ represents the number of data points associated with downstream task $d$, $x_{d,n}$ denotes the embedding for the $n$-th data point in that downstream task, $x_t$ is the embedding of the incoming data sample.

Using this density estimate, we apply null hypothesis significance testing to determine whether $x_t$ belongs to the same distribution as the downstream task data, where the null hypothesis ($H_0$) is ``$x_t$ belongs to the same distribution as the downstream task data" and the alternate hypothesis ($H_a$) is ``$x_t$ does \textbf{not} belong to the same distribution as the downstream task data."
We reject the null hypothesis $H_0$ if the $p$-value of the density estimate $\hat{f}_{\text{vMF-KDE}}(x_t)$ under the assumption of the null hypothesis is less than the significance level $\alpha = 0.05$ and thus consider $x_t$ \textit{irrelevant} to the downstream task.

Since it is difficult to compute the exact $p$-value of the density estimate, we instead approximate it by first computing the kernel density estimate $\hat{f}_{\text{vMF-KDE}}(x_{d, n})$ of each downstream data $x_{d, n}$ and selecting the $0.05$-th quantile value of downstream data density estimates.
$H_0$ is rejected if $\hat{f}_{\text{vMF-KDE}}(x_t)$ is less than the $0.05$-th quantile value of downstream data density estimates.
In short, the incoming data passes the relevance filter for the downstream task $d$ if
\begin{equation}
    \hat{f}_{\text{vMF-KDE}}(x_t) > Q_{0.05} \left ( \{ \hat{f}_{\text{vMF-KDE}}(x_{d, n} ) \}_{n=1}^{N_d} \right )
\end{equation}
where $Q_{0.05}$ is a function that outputs a $0.05$-th quantile value.
This approach enables effective and task-specific data filtering, supporting robust multimodal model training.


\subsection{Specificity Filtering}
\label{subsec:specificity_filtering}
The purpose of specificity filtering is to retain data that is highly informative to the target task, filtering out broader, general data that might introduce irrelevant patterns. 
By focusing on specific data, the model achieves greater data efficiency and faster convergence~\cite{kim2024hype}—key factors in online learning where quick adaptation to essential features can reduce training time and the overall data requirement.

While a prior work, HYPE~\cite{kim2024hype}, employs a hyperbolic multimodal model to measure the specificity of image and text, it is not suitable for online filtering setting since 
(1) it requires a batch of data to assess specificity, (2) incorporating an additional model reduces filtering efficiency, and 
(3) there is currently no hyperbolic model for video-text, which necessitates training a new model from scratch.

In contrast, a more efficient approach would be to reuse the pretrained feature extractor and feature embeddings from previous filtering stages to measure the specificity of incoming data. 
Therefore, we propose using the distance from a root embedding as a simple and effective metric for evaluating specificity.


We define the root embedding $s_r$ as the text embedding of an empty string $r = \text{`` ''}$.
It serves as an anchor for evaluating the specificity of text, as less specific texts tend to be closer to the root embedding than more complicated ones~\cite{alper2024hierarcaps}.
We define the Euclidean distance of any text embedding  $s$  from this root as $\delta_r(s) = ||s - s_r ||$.

The specificity filter then operates by comparing the distance of the incoming data sample's text embedding, $s_t$, to this root against a threshold derived from the downstream task embeddings. 
The incoming data passes the specificity filter for the downstream task $d$ if
\begin{equation}
    \delta_r(s_t) > Q_q \left (\{\delta_r(s_{d,n})\}_{n=1}^{N_d} \right )
\label{eq:specificity_threshold}
\end{equation}
where $Q_q \left ( \{\delta_r(s_{d,n})\}_{n=1}^{N_d} \right )$ denotes the $q$-th quantile of the distances between downstream task text embeddings $s_{d,n}$ and the root $s_r$, which are precomputed prior to the filtering. 
This quantile value serves as a threshold, and we consider a stream sample $s_t$ to be sufficiently specific if Eq.~\ref{eq:specificity_threshold} evaluates to True. 

This specificity filter offers an efficient, task-focused method for assessing specificity, allowing the model to prioritize essential and informative task-relevant data without the need for extensive reconfiguration or an alternate geometric space training.

%% file: sec/3_experiments.tex
\section{Experiments}
\label{sec:experiments}
We present the experimental settings, baseline models, results, and corresponding analysis. 
The central question we address is how our ReSpec-filtered online training contributes to improved performance on the specified downstream tasks.

\subsection{Settings}
\label{subsec:settings}

We evaluate the effectiveness of our online filtering approach by using two real-world noisy source datasets with a foundational model suitable for online training.

\begin{figure*}[t]
    \begin{center}
        \begin{subfigure}{0.48\textwidth}
        \includegraphics[width=\columnwidth]{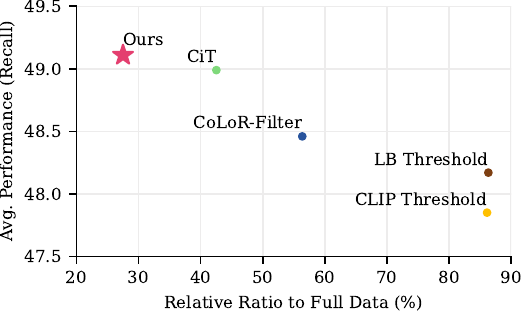}
        \caption{WebVid2M}
    \end{subfigure}
    \hspace{10pt}
    \begin{subfigure}{0.48\textwidth}
        \includegraphics[width=\columnwidth]{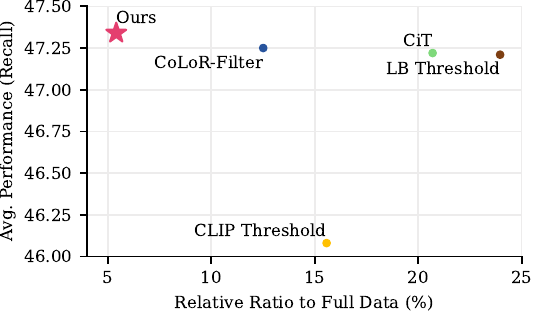}
        \caption{VideoCC3M}
    \end{subfigure}
    \vskip -0.1in
    \caption{\textbf{Performance comparison.}
    We compare our approach to the baselines based on the average performance and the ratio of filtered data size to full data size.
    The average performance is the average of Recall at 1, 5, and 10 across the five downstream tasks.
    Training on full datasets without filtering achieves an average performance of 47.23 on WebVid2M and 42.62 on VideoCC3M.
    }
    \label{fig:top1_perf}
    \end{center}
    \vskip -0.2in
\end{figure*}

\noindent \textbf{Feature extractor. }
LanguageBind~\cite{zhu2024languagebind} is used for extracting video and text feature embeddings for both the downstream and input streaming data.
LanguageBind is a language-based multimodal pre-trained model designed to align five modalities, including video, with language through contrastive learning. 

\noindent \textbf{Noisy source datasets. } 
The input data streams we consider consist of two web-scale noisy video-text datasets.
First, WebVid2M~\cite{bain21webvid} is a large-scale, web-curated dataset consisting of approximately 2.5 million clips and corresponding captions scraped from the stock footage sites.
Another noisy source dataset we use is VideoCC3M~\cite{nagrani2022videocc3m} containing approximately 2.5 million video-text pairs that is constructed by mining videos that are similar to  seed images from CC3M~\cite{sharma2018cc3m}.

\noindent \textbf{Online training model.}
We train a video-language foundation model in an online manner using BT-Adapter~\cite{liu2024btadapter}. 
BT-Adapter enhances pretrained image-language models by introducing temporal modeling capabilities while keeping the original image-language model frozen. 
It achieves this by incorporating a temporal adapter optimized specifically for this purpose. 
This design makes BT-Adapter particularly well-suited for online training settings, as it allows for efficient adaptation, leading to competitive performances with a relatively small amount of data.

\noindent \textbf{Downstream tasks.}
The train set of five text-to-video retrieval downstream tasks are used for the downstream-aware data filtering: MSR-VTT~\cite{xu2016msrvtt}, DiDeMo~\cite{hendricks2017didemo}, LSMDC~\cite{rohrbach2017lsmdc}, ActivityNet~\cite{heilbron2015activitynet}, and YouCook2~\cite{zhou2018youcook2}.
The trained BT-Adapter model is evaluated on the test set of these tasks using Recall at 1, 5, and 10 as evaluation metrics without finetuning.

\subsection{Baselines}
\label{subsec:baselines}
We compare our approach to the following baselines, which include thresholding-based approaches commonly used for data curation and extensions of previous approaches that utilize the downstream task information to the online video-text data filtering.

\noindent \textbf{Cosine similarity thresholding. }
As baselines, we adopt cosine similarity thresholding that is commonly used for data curation~\cite{schuhmann2022laion-5b,schuhmann2021laion-400m,tan2024koala}.
We compute the CLIP~\cite{clip} frame-wise cosine similarity of image and text embeddings and discard data whose cosine similarity is below a certain threshold (CLIP Threshold).
We use three variants of CLIP Threshold, each using the average, max, and middle-frame cosine similarity, and report the best result.
Similarly, we use LanguageBind~\cite{zhu2024languagebind} (LB) to compute the video-text cosine similarity between video and text and apply the threshold values to it (LB Threshold).
We use $\{0.3, 0.28, 0.26, 0.24, 0.22, 0.20\}$ as the threshold values and report the best result.

\noindent \textbf{Online downstream task-aware filtering.}
We adapt previous online downstream task-aware filtering approaches CiT~\cite{xu2023cit} and CoLoR-Filter~\cite{brandfonbrener2024color-filter} to online video-text filtering by using LanguageBind as the filtering network for CiT and as the prior model for CoLoR-Filter.
Since these two approaches do not consider the multimodal alignment between video and text, we use LanguageBind video-text cosine similarity thresholding before applying them for a fair comparison.
We report the best result across the aforementioned video-text cosine similarity threshold values.

\subsection{Results}
Fig.~\ref{fig:top1_perf} reports the results of the online trained BT-Adapter on WebVid2M~\cite{bain21webvid} and VideoCC3M~\cite{sharma2018cc3m}.
We report the average of Recall at 1, 5, and 10 across the five downstream tasks as the average performance (Detailed results are in the Appendix).
On both datasets, using LanguageBind embedding for cosine similarity thresholding performs better than using CLIP.
While two online downstream task-aware filtering baselines, CiT~\cite{xu2023cit} and CoLoR-Filter~\cite{brandfonbrener2024color-filter}, improve performance compared to cosine similarity thresholding (LB Threshold) and further reduce the size of the filtered data, our ReSpec filtering approach not only achieves the best performance but also uses the smallest amount of data across both datasets.

\begin{table}[t]
    \centering
    \resizebox{\columnwidth}{!}{%
    \begin{tabular}{ccccc}
    \toprule
    \textbf{Alignment} & \textbf{Relevance} & \textbf{Specificity} & \textbf{Clip ratio (\%)} & \textbf{Avg. perf.} \\
    \midrule
    \ding{55} & \ding{55} & \ding{55}    &  100\%  &  47.23 \\
    \ding{51} & \ding{55} & \ding{55}   &  86.3\%  &  48.17  \\
    \ding{51} & \ding{51} & \ding{55}   &  29.0\%  &  48.79  \\
    \ding{51} & \ding{51} & \ding{51}   &  \textbf{27.5\%}  &  \textbf{49.11}  \\
    \bottomrule
    \end{tabular}
    }
    \caption{\textbf{Ablation analysis of ReSpec filtering on WebVid2M.} Clip ratio denotes the ratio between the size of filtered data and the size of the entire, unfiltered data. Each component of our approach (alignment, relevance, and specificity) improves the overall performance while reducing the size of filtered data.}
    \label{tab:ablation_analysis}
    \vskip -0.2in
\end{table}

\subsection{Analysis \& Ablations}

\textbf{Ablation on relevance and specificity filtering.}
Tab.~\ref{tab:ablation_analysis} presents the ablation analysis of our filtering framework on WebVid2M, examining the impact of removing individual components (alignment, relevance, and specificity filtering). 
The results indicate that both relevance filtering and specificity filtering reduce the volume of filtered data (and consequently the number of training iterations) while enhancing the overall performance. This supports the importance of considering both task relevance and data specificity for effective and efficient online filtering.

\begin{figure}[t]
    \begin{center}
    \includegraphics[width=\columnwidth]{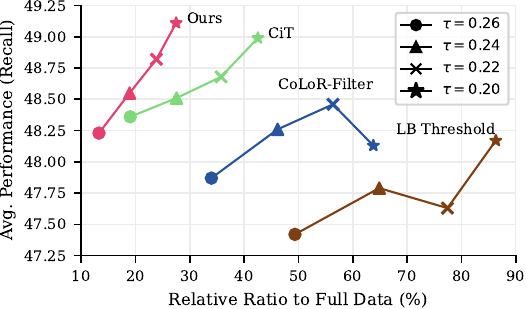}
    \caption{\textbf{Multimodal alignment threshold analysis on WebVid2M.} Our ReSpec filtering achieves the best performance overall across different video-text cosine similiarity threshold values.
    }
    \label{fig:lb_threshold_analysis}
    \end{center}
    \vskip -0.2in
\end{figure}

\noindent
\textbf{Ablation on multimodal alignment threshold.}
Fig.~\ref{fig:lb_threshold_analysis} presents an ablation study and a comparison of different video-text cosine similarity thresholds $\tau$ employed in the multimodal alignment filtering process (as described in \S~\ref{subsec:multimodal_alignment_filtering}), as well as their integration with other baseline methods, including CiT, CoLoR-Filter, and Vanilla LB threshold. 
The results demonstrate that our ReSpec filtering method consistently outperforms or is on par with other baselines on various thresholds, evaluated based on two criteria: average performance and the relative size of the filtered dataset.

\noindent\textbf{Analysis on the relevance filtering modality.}
Relevance filtering (\S~\ref{subsec:relevance_filtering}) is a generic approach that can utilize the video modality, text modality, or a combination of both. 
Tab.~\ref{tab:relevance_modality_ablation} provides an analysis of the impact of different modalities on relevance filtering.

The results indicate that using the video modality alone results in oversampling and poor performance. 
Similarly, the \texttt{Text} $\cup$ \texttt{Video} setting, which requires either \texttt{Text} or \texttt{Video} to be relevant, also suffers from these issues. 
While the \texttt{Text} $\cap$ \texttt{Video} setting, which requires both \texttt{Text} and \texttt{Video} to be relevant, reduces the amount of filtered data, the average performance decreases by 0.17. 

However, by leveraging video embeddings, we can further reduce the number of clips by 26\% with only a marginal influence on average performance. 
This highlights the potential of multimodal approaches, particularly the positive role of video features, in optimizing data selection while maintaining performance. 
These findings suggest that while raw video embeddings may have limitations in modeling the distribution of downstream task data and measuring relevance, careful integration of video features can bring substantial benefits.

To get a better understanding of why text-based relevance filtering outperforms video-based filtering, we look into the concentration parameter $\hat{\kappa}$ from the Eq.~\ref{eq:kappa_estimation} for each downstream task (Tab.~\ref{tab:kappa}).
In most downstream tasks, text embeddings have higher $\hat{\kappa}$ values, indicating that text embeddings are more concentrated, whereas video embeddings are more sparsely distributed.
Although learning on sparsely distributed and diverse data can enhance generalizability~\cite{tirumala2023d4,xu2023youku}, concentrated embeddings with clearer boundaries are better suited for identifying downstream-relevant data, more effective for our filtering objectives.

\begin{table}[t]
    \centering
    \footnotesize
    \begin{tabular}{lcc}
    \toprule
    \textbf{Relevance filtering modality} & \textbf{Clip ratio (\%)} & \textbf{Avg. perf.} \\
    \midrule
    \texttt{Text} & 27.5\% & \textbf{49.11} \\
    \texttt{Video} & 63.7\% & 47.99 \\
    \texttt{Text} $\cup$ \texttt{Video} & 68.5\% & 48.10 \\
    \texttt{Text} $\cap$ \texttt{Video} & \textbf{20.4\%} & 48.94 \\
    \bottomrule
    \end{tabular}
    \vskip -0.05in
    \caption{\textbf{Ablation on the relevance filtering modality.} In the relevance filter, using the text modality outperforms using the video modality or both modalities. \texttt{Text} $\cup$ \texttt{Video} means that the data is considered relevant to the downstream task if either its text or video is considered relevant. 
    \texttt{Text} $\cap$ \texttt{Video} means that data is considered relevant only if both its text and video are considered relevant. The results are from WebVid2M.}
    \label{tab:relevance_modality_ablation}
    \vskip -0.1in
\end{table}

\begin{table}[t]
    \centering
    \footnotesize
    \begin{tabular}{lcc}
    \toprule
    \textbf{Downstream task} & \textbf{Text $\hat{\kappa}$} & \textbf{Video $\hat{\kappa}$} \\
    \midrule
    MSR-VTT~\cite{xu2016msrvtt} & \textbf{693.19} & 509.39 \\
    DiDeMo~\cite{hendricks2017didemo} & \textbf{705.25} & 527.08 \\
    ActivityNet~\cite{heilbron2015activitynet} & \textbf{683.91} & 555.89 \\
    YouCook2~\cite{zhou2018youcook2} & \textbf{1103.50} & 948.60 \\
    LSMDC~\cite{rohrbach2017lsmdc} & 838.17 & \textbf{868.76} \\
    \bottomrule
    \end{tabular}
    \vskip -0.05in
    \caption{\textbf{Estimated value of $\kappa$ for downstream task text and video embeddings.} The estimated value of the concentration parameter $\kappa$ is larger for text embeddings than for video embeddings in most downstream tasks, indicating that the text modality is more concentrated compared to the video modality.}
    \label{tab:kappa}
    \vskip -0.2in
\end{table}

\noindent
\textbf{Ablation on density estimation.}
Relevance filtering (\S~\ref{subsec:relevance_filtering}) employs a non-parametric, data-driven kernel density estimation to estimate the density of incoming data, which is then used for statistical testing to assess its relevance to the downstream task distribution.
An alternative approach is parametric density estimation, assuming the downstream distribution is a single von Mises-Fisher (vMF) distribution. 
This distribution is defined by a mean direction ($\mu$) and a concentration parameter ($\kappa$), estimated via maximum likelihood.
Tab.~\ref{tab:density_estimation_ablation} compares using kernel density estimation to using a single vMF distribution in relevance filtering.
The results show that using a single vMF distribution leads to oversampling, as it less accurately models the downstream task data distribution, resulting in worse performance.

\begin{table}[t]
    \centering
    \footnotesize
    \begin{tabular}{lcc}
    \toprule
    \textbf{Density estimation} & \textbf{Clip ratio (\%)} & \textbf{Avg. perf.} \\
    \midrule
    Kernel density estimation (Ours) & \textbf{27.5\%} & \textbf{49.11} \\
    von Mises-Fisher distribution & 34.1\% & 48.59 \\
    \bottomrule
    \end{tabular}
    \vskip -0.05in
    \caption{\textbf{Ablation on the choice of density estimation on WebVid2M.} In the relevance filter, kernel density estimation samples less data while performing better than modeling each downstream task data using a single von Mises-Fisher distribution. 
    }
    \label{tab:density_estimation_ablation}
    \vskip -0.1in
\end{table}

\noindent
\textbf{Ablation on specificity filter threshold.}
Specificity filtering (\S~\ref{subsec:specificity_filtering}) requires a quantile hyperparameter $q \in [0, 1]$ which is used to determine the task-specific threshold on the distance to the root text embedding.
We provide an ablation analysis on the choice of $q$ in Tab.~\ref{tab:quantile_ablation}.
The results show that while choosing $q=0.1$ achieves best performance, our framework is relatively robust to the choice of $q$.

\begin{table}[t]
    \centering
    \footnotesize
    \begin{tabular}{lcc}
    \toprule
    \textbf{Quantile $q$} & \textbf{Clip ratio (\%)} & \textbf{Avg. perf.} \\
    \midrule
    $q=0.05$ & 28.1\% & 48.55 \\
    $q=0.1$ & 27.5\% & \textbf{49.11} \\
    $q=0.15$ & 27.0\% & 48.94 \\
    $q=0.25$ & 25.8\% & 48.88 \\
    $q=0.50$ & \textbf{21.9\%} & 48.83 \\
    \bottomrule
    \end{tabular}
    \vskip -0.05in
    \caption{\textbf{Ablation on the choice of quantile $q$.} While choosing $q=0.1$ shows the best performance, specificity filtering is relatively robust to the choice of $q$. The results are from WebVid2M.}
    \label{tab:quantile_ablation}
    \vskip -0.1in
\end{table}

\noindent
\textbf{Analysis on the filtered data.}
To measure how distributionally similar our filtered data is to the downstream task data, we measure Fr\'echet distance between the LanguageBind embeddings of the filtered data and that of downstream task data.
To consider both video and language aspects, we use concatenation of video and text embeddings.
We also measure the KL divergence between downstream task texts and the filtered texts using hashed N-gram representation as in DSIR~\cite{xie23dsir}.
The results in Tab.~\ref{tab:frechet_distance} show that while two downstream task-aware filtering baselines (CiT and CoLoR-Filter) reduce the Fr\'echet distance and the text KL divergence to the downstream task data, the filtered data using our approach is the closest to the downstream task data.
We conjecture that this is because our relevance filtering explicitly considers the distributional aspect of each downstream task using density estimation and statistical testing, while CiT and CoLoR-Filter do not.

\begin{table}[t]
    \centering
    \footnotesize
    \begin{tabular}{lcc}
    \toprule
    \textbf{Filtering method} & \textbf{Fr\'echet distance} & \textbf{Text KL divergence}\\
    \midrule
    ReSpec (Ours) & \textbf{0.2371} & \textbf{0.4371} \\
    CiT~\cite{xu2023cit} & 0.2513  & 0.4488 \\
    CoLoR-Filter~\cite{brandfonbrener2024color-filter} & 0.2977 & 0.4921 \\
    LB Threshold & 0.3028 & 0.5035 \\
    \bottomrule
    \end{tabular}
    \vskip -0.05in
    \caption{\textbf{Fr\'echet distance using LanguageBind embeddings and text KL divergence using hashed N-gram features~\cite{xie23dsir} on WebVid2M.} While both CiT and CoLoR-Filter reduce the Fr\'echet distance and KL divergence to the downstream tasks data, the filtered data using our approach is distributionally closest to it.}
    \label{tab:frechet_distance}
    \vskip -0.1in
\end{table}

\noindent
\textbf{Analysis on the filtering efficiency.}
We compare the computational cost of our approach to other baselines in Tab.~\ref{tab:computation_cost}.
We assume a backward pass is twice as expensive as a forward pass~\cite{brandfonbrener2024color-filter}.
LB Threshold only requires a single forward pass of LanguageBind~\cite{zhu2024languagebind} for each incoming data ($PN$) and computing the dot product between video and text embeddings ($N$), being the most efficient.
Our approach incurs a one-time prior cost of computing embeddings of downstream task data ($PM$) and computing their distance to the root text embedding ($M$) as well as additional inference cost for computing dot product between incoming data embedding and downstream task embeddings ($MN$) and computing distance to the root embedding ($N$).
In addition to the inference cost for computing the dot product ($MN$), CiT entails the cost of parameter update using the filtered data ($2\gamma PN$) and requires recomputing the downstream task embeddings whenever the filtering model parameter is updated ($\gamma PMN/b$).
CoLoR-Filter involves the prior cost of finetuning a prior model ($3PM$) and needs two forward passes and crossmodal dot product computation for each incoming data ($2PN + 2N$).
Since the computation cost of LanguageBind forward pass is larger than the computation cost of computing the dot product between incoming data embedding and downstream task embeddings (\ie, $P > M$), our approach is more computationally efficient than CiT and CoLoR-Filter.

\begin{table}[t]
    \centering
    \resizebox{\columnwidth}{!}{%
    \begin{tabular}{lccc}
    \toprule
    \textbf{Filtering method} & \textbf{Prior} & \textbf{Inference} & \textbf{Param. update} \\
    \midrule
    LB Threshold & $0$ & $PN + N$ & $0$ \\
    ReSpec (Ours) & $PM + M$ & $PN + MN + 2N$ & $0$ \\
    CiT~\cite{xu2023cit} & $0$ & $PN + MN + N + \gamma PMN / b$ & $2\gamma PN$ \\
    CoLoR-Filter~\cite{brandfonbrener2024color-filter} & $3PM$ & $2PN + 2N$ & 0 \\
    \bottomrule
    \end{tabular}
    }
    \vskip -0.05in
    \caption{\textbf{Computation cost comparison of filtering methods.}
    The column \textbf{Prior} indicates the cost incurred before filtering, while \textbf{Inference} represents the total forward pass cost during filtering. 
    Here, $M$ is the size of the downstream task data, $N$ is the size of the streaming dataset (\eg, WebVid2M), and $P$ denotes the inference cost of LanguageBind~\cite{zhu2024languagebind}. 
    $\gamma$ represents the ratio of filtered data, and $b$ is the batch size used for parameter updates in CiT. 
    Note, CiT additionally incurs \textbf{Parameter update} costs due to occasional updates to the filtering model using the filtered data.
    }
    \label{tab:computation_cost}
    \vskip -0.1in
\end{table}

%% file: sec/4_conclusion.tex
\section{Conclusion}
\label{sec:conclusion}
The increasing volume of multimodal video-text data demands approaches to address storage and computational constraints, especially in online learning scenarios. 
To meet these demands, we proposed ReSpec, an efficient online filtering framework designed to prioritize data based on modality alignment, task relevance, and specificity. 
By leveraging these criteria, ReSpec effectively selects high-value data for the target downstream tasks without the need for extensive storage or high-latency processing. 
Evaluated on large-scale datasets, ReSpec demonstrated its capability to achieve state-of-the-art performance in zero-shot video retrieval tasks while significantly reducing the data requirement. 
This work highlights the potential of real-time data filtering for enabling efficient, adaptive learning in real-world, resource-constrained environments.

%% file: sec/5_acknowledgments.tex
\section{Acknowledgments}
\label{sec:acknowledgments}
This work was supported by LG AI Research, The Allen Institute for Artificial Intelligence, 
Institute of Information \& Communications Technology Planning \& Evaluation
(IITP) grant funded by the Korea government (MSIT) (No. RS-2019-II191082, SW StarLab), 
grant funded by the Korea government (MSIT) (No.~RS-2021-II211343, Artificial Intelligence Graduate School Program (Seoul National University)),
ITRC(Information Technology Research Center) grant funded by the Korea government (Ministry of Science and ICT)(IITP-2025-RS-2024-00437633).

%% file: sec/6_supplementary.tex
\clearpage
\setcounter{page}{1}
\maketitlesupplementary

\section{Appendix: Table of Contents}

The Appendix enlists the following additional materials. 
\begin{enumerate}[label=\Roman*.]
    \item Extended Related Works. \S~\ref{sec:extended_related_works}
     \begin{enumerate}[label=\roman*.]
            \item Robust Learning on Noisy Datasets \S~\ref{subsec:robust_learning}
            \item Online Learning \S~\ref{subsec:online_learning}
        \end{enumerate}
    \item Experiment Setting. \S~\ref{sec:experiment_details}
    \begin{enumerate}[label=\roman*.]
            \item Dataset Details \S~\ref{subsec:dataset_details}
            \item Baseline Details \S~\ref{subsec:baseline_details}
            \item Training Details \S~\ref{subsec:training_details}
        \end{enumerate}
    \item Extended Results. \S~\ref{sec:extended_results}
    \begin{enumerate}[label=\roman*.]
            \item Additional Dataset Results  \S~\ref{subsec:add_dataset_results}
            \item Additional Architecture Results  \S~\ref{subsec:add_architect_results}
            \item Generalization to Online Image-Text Filtering  \S~\ref{subsec:imagetext_results}
            \item Multi-Dataset Training Results  \S~\ref{subsec:multi_dataset_training}
            \item Per-Task Results \S~\ref{subsec:per_task}
        \end{enumerate}
    \item Extended Analysis. \S~\ref{sec:extended_anal}
    \begin{enumerate}[label=\roman*.]
        \item Embedding Robustness \S~\ref{subsec:embedding_robustness}
        \item Relevance Filter Ablations \S~\ref{subsec:relevance_ablations}
        \item Baseline Ablations \S~\ref{subsec:baseline_ablations}
        \item Qualitative Analysis \S~\ref{subsec:qualitative_analysis}
    \end{enumerate}

\end{enumerate}

\section{Extended Related Works}
\label{sec:extended_related_works}

\subsection{Robust Learning on Noisy Dataset} \label{subsec:robust_learning}

Noisy correspondence learning focuses on research related to mismatched pairs in multimodal data. 
Its goal is to train models robustly in noisy correspondences, generally by measuring the degree of alignment between modalities and reflecting it in the training process.
In early research (NCR~\cite{huang2021learning}), the clean/noisy data is classified using a GMM-based method by utilizing the loss difference between clean and noisy data, which stems from co-training and memorization effects. 
NPC~\cite{zhang2024negative} measures the correspondence based on the performance difference between the current data and clean data trained on a similar sample.

This approach has recently been extended to video-text datasets, where temporal misalignment is considered. 
In such cases, frame-text alignment is measured using Optimal Transport (NorTon~\cite{lin2024multi}) or mutual agreement-based methods (TempCLR~\cite{yang2023tempclr}), which are then reflected in the learning process.
Noisy correspondence learning generally involve storing the entire dataset and conducting several iterations/epochs to train a generalizable model from the noisy dataset, incurring significant storage and computational costs.

\subsection{Online Learning}
\label{subsec:online_learning}

Online learning has largely been explored in the context of continual learning, which studies the problem of learning sequential tasks while alleviating catastrophic forgetting on the past tasks~\cite{mai2022online,aljundi2019gradient,wang2022ocl-nds,cai2021cvt}.
OCL-NDS~\cite{wang2022ocl-nds} proposes adaptive learning rate scheduling and replay buffer size adaptation algorithm for online continual learning with natural distribution shift.
CVT~\cite{cai2021cvt} proposes an attention-based mechanism to mitigate the catastrophic forgetting for online continual contrastive learning.

Online learning has also been used to improve the training efficiency by specializing a model on specific target tasks~\cite{mullapudi2019omd,boldo2023query,farhadi2020enabling}.
For instance, OMD~\cite{mullapudi2019omd} trains a small network on a long video stream using online model distillation from high capacity teacher model, obtaining specialized model that performs on par with the much larger teacher on the target video stream.

\section{Experiment Details}
\label{sec:experiment_details}

\subsection{Dataset Details}
\label{subsec:dataset_details}

The input data stream we consider consists of two web-scale noisy video-text datasets.
First, VideoCC3M~\cite{nagrani2022videocc3m} is a large-scale, web-curated dataset containing approximately 2.5 million video-text pairs.
It is constructed by mining videos that are similar to seed images from CC3M~\cite{sharma2018cc3m} and transferring the images' captions to the videos.
Since both the construction of CC3M and the video mining of VideoCC3M are done automatically, the videos and texts in VideoCC3M are considered weakly paired and contains various types and degrees of noise.

Another noisy source dataset we use is WebVid2M~\cite{bain21webvid}, consisting of approximately 2.5 million clips and corresponding captions scraped from the stock footage sites.
While WebVid2M is considered more \textit{clean} than VideoCC3M since the captions are human-generated, it still contains noisy correspondences between video and text, such as text containing the metadata of the video rather than its description.

\subsection{Baseline Details}
\label{subsec:baseline_details}

\noindent \textbf{Cosine similarity thresholding. }
When constructing a large-scale multimodal datasets, it has become a common practice to compute CLIP~\cite{clip} cosine similarity of image and text embeddings and drop samples below a certain threshold to filter out unsuitable image-text pairs.
For instance, the threshold of $0.3$ is used in LAION-400M~\cite{schuhmann2021laion-400m} and the thresholds of $0.28$ and $0.26$ are used in LAION-5B~\cite{schuhmann2022laion-5b} depending on the language of the text.
Similar approach is also used to filter video-text datasets, as Koala~\cite{tan2024koala} computes the frame-wise CLIP cosine similarity and only use samples whose max frame-wise cosine similarity is greater than or equal to $0.26$ as training data.
Thus as baselines, we compute frame-wise image-text CLIP cosine similarity and apply threshold values of $0.3, 0.28, 0.26, 0.24, 0.22$, and $0.2$ on the cosine similarities.
We use three variants of CLIP Threshold, each using the average, max, and middle-frame cosine similarity.
Similarly, we also compute video-text cosine similarity between text and video using LanguageBind~\cite{zhu2024languagebind} and apply the threshold values on.
Note that these baselines only consider the multimodal alignment and do not account for the downstream task relevance and specificity.

\noindent \textbf{Online downstream task-aware filtering.}
We adapt previous online downstream task-aware filtering approaches CiT~\cite{xu2023cit} and CoLoR-Filter~\cite{brandfonbrener2024color-filter} to our setting.

While CiT~\cite{xu2023cit} is originally designed for training a single joint filtering and training network, we adapt it to our setting by using separate filtering (LanguageBind) and training (BT-Adapter) networks for fair comparison.
CiT uses cosine similarity between text of incoming data and texts from downstream task data as a measure of relevance and selects sample whose cosine similarity to downstream task data is larger than a certain threshold.
The selected samples are also used for the training of the filtering network.
For computational efficiency, we only update the parameters of the projection layers.
Following the original work, we use AdamW~\cite{loshchilov2018adamw} optimizer with learning rate of 5e-4 and weight decay of $1.0$, and batch size of 1,536 (CiT single GPU setting) for the filtering network parameter update.
We use the text cosine similarity threshold of $0.55$.
Section~\ref{subsec:baseline_ablations} provide ablation on the text cosine similarity threshold.

CoLoR-Filter~\cite{brandfonbrener2024color-filter} is originally designed to select downstream relevant data for language models by finetuning the prior language model on downstream dataset and using the difference of language modeling loss (negative log probability) of the data computed using the finetuned model and the prior model as the criterion.
We extend it to online sample-wise video-text data filtering by using LanguageBind as the prior model and finetuning it on downstream datasets via contrastive loss.
For copmutational efficiency, we finetune the projection layers of LanguageBind using the downstream datasets for ten epochs using AdamW optimizer with learning rate of 1e-3 and weight decay of $0.2$, and batch size of 4096.

While the original CoLoR-Filter ranks samples within a mini-batch and select certain proportion of the data based on the ranking, we modify it to work in online, sample-wise setting.
We use the difference of the video-text cosine similarity of the data between the finetuned and the prior model as the criterion, and only sample when the cosine similarity measured using the finetuned model is larger.
We also provide the result of using the original ranking-based selection using mini-batch of data in Section~\ref{subsec:baseline_ablations}.

Since these two approaches do not consider the multimodal alignment between video and text (\ie cleanness), we use LanguageBind video-text cosine similarity thresholding before applying them for fair comparison.
We report the best result across the aforementioned video-text cosine similarity threshold values.

\subsection{Training Details}
\label{subsec:training_details}

For ReSpec and other baselines, we use the filtered data for the online training of BT-Adapter~\cite{liu2024btadapter}.
We use BT-Adapter with OpenAI CLIP-L/14 backbone, and follow the implementation details of the original work~\cite{liu2024btadapter} for the masking ratio, temperature scale, and the number of adapted layers.
We use AdamW~\cite{loshchilov2018adamw} optimizer with learning rate 2e-6 and weight decay 0.05 as in the original implementaion, with batch size of 52 for WebVid2M and 100 for VideoCC3M. 

\begin{table*}[t]
    \centering
    \begin{footnotesize}
    \setlength{\tabcolsep}{2.5pt}
    \resizebox{\textwidth}{!}{%
    \begin{tabular}{lccccccccccccccccccc}
    \toprule
    \textbf{Model} & \textbf{Clip ratio (\%)} & \multicolumn{3}{c}{\textbf{MSR-VTT}} & \multicolumn{3}{c}{\textbf{DiDeMo}} & \multicolumn{3}{c}{\textbf{ActivityNet}} & 
    \multicolumn{3}{c}{\textbf{YouCook2}} & 
    \multicolumn{3}{c}{\textbf{LSMDC}} & \textbf{Avg. Perf.} \\
    \midrule
     &  & \textbf{R1} & \textbf{R5} & \textbf{R10} & \textbf{R1} & \textbf{R5} & \textbf{R10} & \textbf{R1} & \textbf{R5} & \textbf{R10} & 
     \textbf{R1} & \textbf{R5} & \textbf{R10} & 
     \textbf{R1} & \textbf{R5} & \textbf{R10} & \\
    \midrule
    Full data               &100.00\% & 39.70 & 64.70 & 72.80 & 34.52 & 60.02 & 69.64 & 38.45 & 68.52 & 80.42 & 11.07 & 26.29 & 36.22 & 20.70 & 39.30 & 46.10 & 47.23\\
    CLIP Avg Threshold      & 94.51\% & 41.50 & 64.60 & 73.80 & 35.62 & 59.42 & 70.24 & 38.89 & 68.01 & 80.17 & 10.70 & 27.26 & 37.19 & 22.88 & 40.36 & 46.25 & 47.79\\
    CLIP Mid Threshold      & 86.11\% & 40.60 & 64.50 & 73.00 & 35.42 & 61.31 & 70.44 & 39.23 & 68.79 & 80.49 & 10.70 & 26.70 & 36.43 & 23.20 & 40.40 & 46.60 & 47.85\\
    CLIP Max Threshold      & 98.11\% & 41.30 & 63.70 & 72.50 & 35.42 & 60.71 & 69.94 & 38.67 & 68.50 & 80.35 & 10.79 & 27.50 & 36.72 & 22.00 & 41.50 & 47.40 & 47.80\\
    LB Threshold            & 86.32\% & 40.20 & 64.80 & 73.30 & 36.08 & 61.15 & 70.76 & 40.17 & 69.52 & 81.25 & 10.88 & 27.25 & 37.03 & 22.50 & 40.50 & 47.10 & 48.17\\
    CiT~\cite{xu2023cit}    & 42.53\% & 42.00 & 66.00 & 75.60 & 35.84 & 61.78 & 70.79 & 40.57 & 70.15 & 82.09 & 10.76 & 27.96 & 37.38 & 24.00 & 41.20 & 48.70 & 48.99\\
    CoLoR-Filter~\cite{brandfonbrener2024color-filter} & 56.36\% & 41.70 & 65.50 & 73.30 & 34.82 & 61.81 & 71.33 & 40.50 & 69.21 & 80.82 & 11.02 & 27.56 & 37.24 & 23.40 & 40.90 & 47.80 & 48.46\\
    \textbf{ReSpec (ours)}  & 27.50\% & 42.10 & 67.00 & 76.10 & 36.31 & 62.30 & 72.42 & 40.50 & 69.79 & 81.69 & 11.33 & 26.75 & 37.20 & 24.10 & 40.60 & 48.50 & 49.11\\ 

    \bottomrule
    \end{tabular}
    }
    \end{footnotesize}
    \caption{\textbf{Performance on 5 downstream tasks trained with WebVid2M dataset.} }
    \label{tab:btadapter_webvid}
\end{table*}

\begin{table*}[t]
    \centering
    \begin{footnotesize}
    \setlength{\tabcolsep}{2.5pt}
    \resizebox{\textwidth}{!}{%
    \begin{tabular}{lcccccccccccccccccc}
    \toprule
    \textbf{Model} & \textbf{Clip ratio (\%)} & \multicolumn{3}{c}{\textbf{MSR-VTT}} & \multicolumn{3}{c}{\textbf{DiDeMo}} & \multicolumn{3}{c}{\textbf{ActivityNet}} & 
    \multicolumn{3}{c}{\textbf{YouCook2}} & 
    \multicolumn{3}{c}{\textbf{LSMDC}} & \textbf{Avg. Perf.} \\
    \midrule
     &  & \textbf{R1} & \textbf{R5} & \textbf{R10} & \textbf{R1} & \textbf{R5} & \textbf{R10} & \textbf{R1} & \textbf{R5} & \textbf{R10} & 
     \textbf{R1} & \textbf{R5} & \textbf{R10} & 
     \textbf{R1} & \textbf{R5} & \textbf{R10} & \\
    \midrule
    Full data               & 100.00\% & 35.00 & 58.80 & 69.80 & 34.79 & 58.97 & 70.27 & 31.70 & 60.48 & 74.61 & 6.86 & 18.33 & 25.62 & 18.90 & 34.10 & 41.10 & 42.62\\
    CLIP Avg Threshold      & 15.57\% & 39.60 & 64.70 & 73.70 & 34.72 & 60.02 & 68.65 & 35.34 & 65.10 & 77.56 & 9.59 & 23.76 & 33.17 & 20.90 & 38.40 & 46.00 & 46.08\\
    CLIP Mid Threshold      & 43.62\% & 39.90 & 63.50 & 73.60 & 35.32 & 59.72 & 70.04 & 34.87 & 62.83 & 76.38 & 9.24 & 23.18 & 32.54 & 22.10 & 38.10 & 44.20 & 45.70\\
    CLIP Max Threshold      & 39.36\% & 38.30 & 64.60 & 74.40 & 34.79 & 59.86 & 70.37 & 35.37 & 63.67 & 76.58 & 9.07 & 23.55 & 32.57 & 21.70 & 38.80 & 45.50 & 45.94\\
    LB Threshold            & 23.95\% & 40.00 & 65.90 & 75.40 & 35.74 & 60.79 & 70.89 & 37.33 & 66.76 & 79.06 & 9.99 & 24.24 & 33.43 & 22.48 & 39.36 & 46.85 & 47.21\\
    CiT                     & 20.69\% & 40.60 & 65.90 & 75.70 & 35.64 & 62.18 & 70.89 & 37.78 & 65.77 & 78.37 & 10.01 & 24.89 & 33.10 & 21.98 & 38.96 & 46.55 & 47.22\\
    CoLoR-Filter            & 12.51\% & 41.30 & 65.10 & 76.10 & 35.45 & 60.40 & 70.00 & 37.87 & 66.58 & 79.44 & 10.10 & 24.96 & 34.23 & 22.70 & 38.50 & 46.10 & 47.25\\
    \textbf{ReSpec (ours)}  & 5.41\% & 40.80 & 65.50 & 75.00 & 34.99 & 61.45 & 70.17 & 37.28 & 66.88 & 79.78 & 10.27 & 25.83 & 35.01 & 21.68 & 39.06 & 46.45 & 47.34\\
    \bottomrule
    \end{tabular}
    }
    \end{footnotesize}
    \caption{\textbf{Performance on 5 downstream tasks trained with VideoCC3M dataset.}}
    \label{tab:btadapter_cc3m}
\end{table*}

\section{Extended Results}
\label{sec:extended_results}

\begin{figure}[t]
    \begin{center}
    \includegraphics[width=\columnwidth]{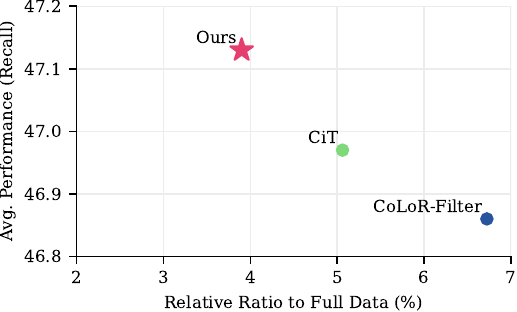}
    \caption{\textbf{Performance comparison on HowTo10M.}
    We compare our approach to the top performing baselines based on the average performance and the ratio of filtered data size to full data size (HowTo10M here).
    The average performance is the average of Recall at 1, 5 and 10 across five downstream tasks.
    }
    \label{fig:top1_bt_howto}
    \end{center}
\end{figure}

\begin{figure}[th]
    \begin{center}
    \includegraphics[width=\columnwidth]{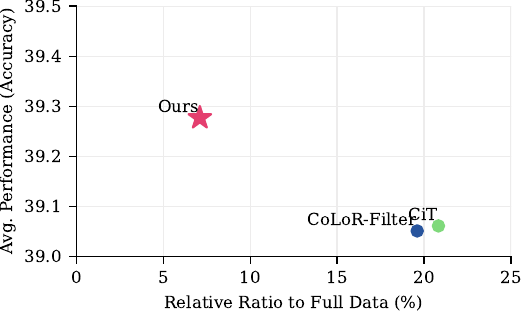}
    \caption{\textbf{Additional architecture performance comparison on VideoCC3M}
    We compare our approach using the FrozenBiLM architecture~\cite{yang2022frozenbilm} to the top performing baselines based on the average performance and the ratio of filtered data size to full data size (VideoCC3M here).
    The average performance is the average of Recall at 1, 5 and 10 across four downstream tasks.
    }
    \label{fig:top1_bilm_cc3m}
    \end{center}
\end{figure}


\begin{figure}[t]
    \begin{center}
        \begin{subfigure}{\columnwidth}
        \includegraphics[width=\columnwidth]{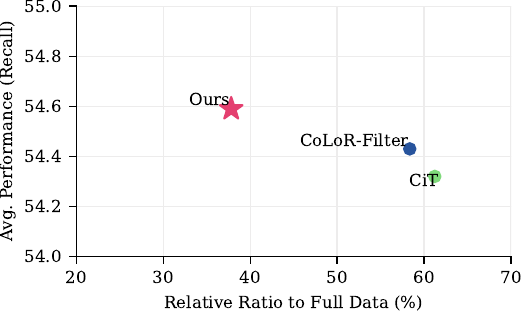}
        \caption{CC3M}
        \end{subfigure}

        \vspace{5pt} 

        \begin{subfigure}{\columnwidth}
        \includegraphics[width=\columnwidth]{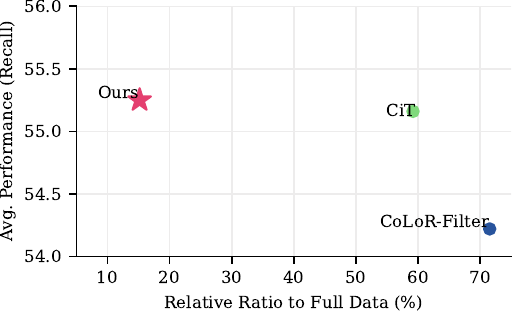}
        \caption{CC12M}
        \end{subfigure}

        \vspace{5pt} 

        \begin{subfigure}{\columnwidth}
        \includegraphics[width=\columnwidth]{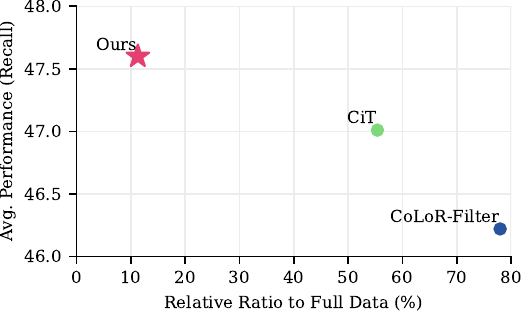}
        \caption{LAION10M}
        \end{subfigure}

        \caption{\textbf{Performance comparison on online image-text filtering} We compare our approach to the baselines based on two metrics: average performance and the ratio of filtered data size to full data size. The evaluation is conducted on three datasets: CC3M~\cite{sharma2018cc3m}, CC12M~\cite{changpinyo2021cc12m}, and a 10M subset of LAION-400M~\cite{schuhmann2021laion-400m}. The average performance is computed as the mean of Recall at 1, 5, and 10 across two downstream tasks.}
    \label{fig:top1_image_text}
    \end{center}
\end{figure}

\subsection{Additional Dataset Results}
\label{subsec:add_dataset_results}
We conduct additional experiments on the HowTo10M dataset, a subset of HowTo100M~\cite{miech2019howto100m} comprising approximately 10\% of the original data. 
Unlike the previously used datasets, HowTo10M is a web-crawled video-text dataset relying on Automatic Speech Recognition (ASR) for video-text alignment. 
This experiment additionally investigates whether our model maintains its performance on an ASR-based video-text dataset.

As shown in Fig.~\ref{fig:top1_bt_howto}, ReSpec delivers the best performance, requiring the least amount of data while achieving the highest average performance across five downstream tasks.

\subsection{Additional Architecture and Downstream Tasks Results}
\label{subsec:add_architect_results}
In addition to BT-Adapter~\cite{liu2024btadapter}, we also validate the efficacy of ReSpec on another architecture, FrozenBiLM~\cite{yang2022frozenbilm}, which is also well-suited for online training.
FrozenBiLM endows the pretrained bi-directional language model with zero-shot video question answering capabilities by freezing the visual backbone and the bi-directional language model and training visual-text projection and adapter layers using the masked language modeling (MLM) objective.

Unlike BT-Adapter, which used zero-shot video-text retrieval as downstream tasks, 
we use four zero-shot open-ended video question-answering tasks (MSRVTT-QA~\cite{xu2017msrvtt-msvd-qa}, MSVD-QA~\cite{xu2017msrvtt-msvd-qa}, ActivityNet-QA~\cite{yu2019activitynet-qa}, and TGIF-QA~\cite{jang2017tgif-qa}) and one zero-shot video-conditioned fill-in-the-blank task (LSMDC-FIB~\cite{maharaj2017lsmdc-fib}) as downstream tasks of FrozenBiLM.

Fig.~\ref{fig:top1_bilm_cc3m} reports the result of the online trained FrozenBiLM on video-text data filtered from VideoCC3M.
ReSpec outperforms CiT and CoLoR-Filter in terms of average performance (average of top-1 and top-10 accuracy) while being most efficient in terms of the amount of the selected data.

\subsection{Generalization to Online Image-Text Filtering}
\label{subsec:imagetext_results}

While we mainly focus on the \textit{video}-text domain since online filtering is more critical for video data, where storage and computational demands are significantly higher, ReSpec is generalizable to online image-text filtering.
To demonstrate the generalizability, we conduct experiments on online image-text filtering and training by using CLIP ViT-B/16 features for filtering and training CiT~\cite{xu2023cit} architecture with ViT-B/16 image encoder pretrained on ImageNet21k and pretrained SimCSE-BERT$_\text{base}$ text encoder using the filtered image-text pairs in an online manner.
we freeze the pretrained image encoder and train the text encoder and two projection layers.
we use COCO~\cite{lin2014microsoft} and Flickr30k~\cite{young2014image} retrieval as the downstream tasks.

Fig.~\ref{fig:top1_image_text} compares ReSpec to CiT and CoLoR-Filter when using CC3M~\cite{sharma2018cc3m}, CC12M~\cite{changpinyo2021cc12m}, and 10M subset of Laion-400M~\cite{schuhmann2021laion-400m} as the noisy source datasets.
in all three settings, ReSpec achieves the best average performance (average of Recall at 1, 5, and 10) while using the smallest amount of data.

\subsection{Multi-Dataset Training Results}
\label{subsec:multi_dataset_training}
We conduct two multi-dataset training experiments: one using a sequential stream from VideoCC3M to WebVid2M, and the other with a randomly shuffled stream of both datasets. 
We evaluate performance using two metrics: average performance (mean Recall at 1, 5, and 10 across five downstream tasks) and the relative ratio of filtered data size to total data size.

\begin{figure}[t]
    \begin{center}
    \includegraphics[width=\columnwidth]{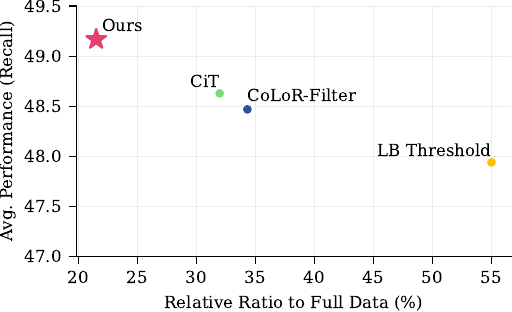}
    \caption{\textbf{Online multi-dataset (VideoCC3M $\rightarrow$ WebVid2M) filtering and training performance comparison.}
    We evaluate our approach using the FrozenBiLM architecture~\cite{yang2022frozenbilm} and compare it against the top-performing baselines based on two key metrics: average performance and the ratio of filtered data size to full data size (CC3M in this case). 
    The average performance is calculated as the mean of the Recall at 1, 5, and 10 across four distinct downstream tasks.
    }

    \label{fig:top1_mutlidataset}
    \end{center}
\end{figure}

First, Fig.~\ref{fig:top1_mutlidataset} compares sequential multi-dataset training (VideoCC3M → WebVid2M) across different approaches. 
Our method outperforms CiT in average performance and requires significantly less data, with a ratio of 21\%, compared to CiT's 31\%. 
In contrast, CoLoR-Filter shows slightly lower performance and a data usage ratio similar to CiT's, while the LB Threshold method requires the largest data volume (55\%) but performs the worst. 
This highlights our approach's efficiency in minimizing data requirements while maintaining top-tier recall performance across downstream tasks with sequential multi-dataset training.

\begin{figure}[t]
    \begin{center}
    \includegraphics[width=\columnwidth]{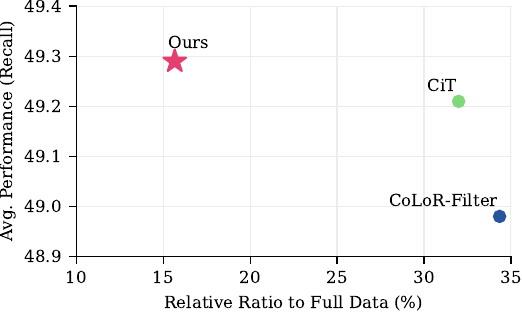}
    \caption{\textbf{Online multi-dataset (randomly shuffled VideoCC3M + WebVid2M) filtering and training performance comparison.}
    We compare our approach to the top performing baselines based on the average performance and the ratio of filtered data size to full data size (VideoCC3M + WebVid2M here).
    The average performance is the average of Recall at 1, 5 and 10 across five downstream tasks.
    }
    \label{fig:top1_bt_mixed}
    \end{center}
\end{figure}

Second, Fig.~\ref{fig:top1_bt_mixed} shows results from training on a randomly shuffled stream of VideoCC3M and WebVid2M. 
This experiment tests whether the model retains robust performance despite the random dataset order. Our approach again outperforms others, requiring only 16\% of the data, compared to CiT's 32\% and CoLoR-Filter's 34\%. 
This further underscores the efficiency of our method in minimizing data usage while maintaining state-of-the-art recall performance across downstream tasks, even with randomly shuffled multi-datasets.

\begin{figure*}[t]
    \begin{center}
        \begin{subfigure}{0.30\textwidth}
        \includegraphics[width=\columnwidth]{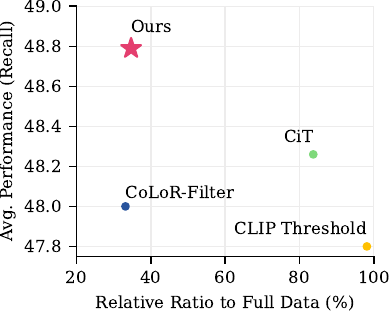}
        \caption{Using CLIP Max embedding}
    \end{subfigure}
    \hspace{5pt}
    \begin{subfigure}{0.30\textwidth}
        \includegraphics[width=\columnwidth]{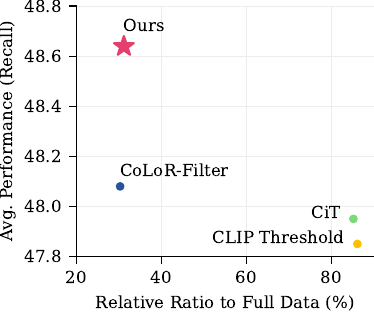}
        \caption{Using CLIP Mid embedding}
    \end{subfigure}
    \hspace{5pt}
    \begin{subfigure}{0.30\textwidth}
        \includegraphics[width=\columnwidth]{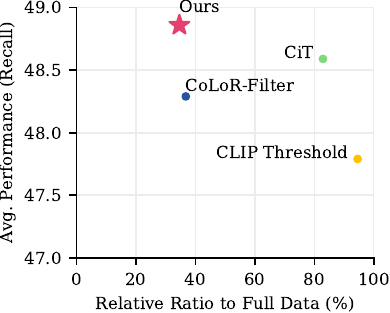}
        \caption{Using CLIP Avg embedding}
    \end{subfigure}
    \caption{\textbf{Performance comparison using Max/Mid/Avg CLIP embeddings}
    We compare our approach to the baselines based on the average performance and the ratio of filtered data size to full data size.
    The average performance is the average of Recall at 1, 5, and 10 across the five downstream tasks.
    All experiments shown are based on WebVid2M.
    }
    \label{fig:top1_clipembed_perf}
    \end{center}
    \vskip -0.2in
\end{figure*}

\subsection{Per-Task Results}
\label{subsec:per_task}

Tab.~\ref{tab:btadapter_webvid}--\ref{tab:btadapter_cc3m} shows the recall at 1, 5, and 10 for each downstream tasks and the average of the recalls across five downstream tasks (Avg. Perf.), along with the clip ratio (the ratio of the filtered data size over the original dataset size).
ReSpec achieves the best average performance while filtering the least amount of data in both WebVid2M and VideoCC3M.


\section{Extended Analysis}
\label{sec:extended_anal}

\subsection{Embedding Robustness Analysis}
\label{subsec:embedding_robustness}
Fig.~\ref{fig:top1_clipembed_perf} presents a comparative evaluation of various baseline methods using three distinct variants of CLIP embeddings: Max, Mid, and Avg. 
Across all embedding types, our proposed method consistently outperforms competing approaches, achieving the highest average recall while making use of small data subsets for training.

These results show the robustness of our approach, demonstrating its effectiveness across diverse embedding variants, in addition to the primary LanguageBind embeddings used in our main experiments. 
This consistent performance further highlights the versatility of the proposed method, emphasizing its potential for broad applicability across a range of tasks and datasets.

\begin{table}[t]
    \centering
    \footnotesize
    \begin{tabular}{lcc}
    \toprule
    \textbf{Relevance Filter} & \textbf{Clip ratio (\%)} & \textbf{Avg. perf.} \\
    \midrule
    Cosine similarity ($\text{threshold}=0.55$)           & 49.99\% & 48.55 \\
    Gaussian distribution modeling     & 45.51\% & 48.23 \\
    vMF distribution modeling          & 34.12\% & 48.95 \\
    vMF kernel density estimation (ours) & \textbf{27.50\%} & \textbf{49.11} \\
    \bottomrule
    \end{tabular}
    \vskip -0.1in
    \caption{\textbf{Ablation of relevance filters on WebVid2M}
    }
    \label{tab:ablate_relevance_filter}
\end{table}

\subsection{Relevance Filter Ablations} 
\label{subsec:relevance_ablations}
To better understand and justify the design choice of our relevance filter, we conduct three comparisons. 
The first comparison involves using cosine similarity on downstream data, the second compares modeling each downstream task using a Gaussian distribution, and the third explores the use of a von Mises-Fisher (vMF) distribution to model each downstream tasks. 
As shown in Table \ref{tab:ablate_relevance_filter}, our relevance filter with vMF kernel density estimation outperforms the other approaches by a significant margin while requiring the least amount of data.

\subsection{Baseline Ablations}
\label{subsec:baseline_ablations}

CiT~\cite{xu2023cit} requires the hyperparameter $\tau_\text{text}$ that is used as a threshold on the cosine similarity between incoming data and downstream task text embeddings.
Tab.~\ref{tab:cit_threshold_ablation} shows the ablation on the choice of $\tau_\text{text}$.
As in the main experimental results of the original CiT paper, choosing $\tau_\text{text}=0.55$ shows the best performance.

\begin{table}[t]
    \centering
    \footnotesize
    \begin{tabular}{lcc}
    \toprule
    \textbf{Threshold $\tau_\text{text}$} & \textbf{Clip ratio (\%)} & \textbf{Avg. perf.} \\
    \midrule
    $\tau_\text{text}=0.5$ & 57.4\% & 48.58 \\
    $\tau_\text{text}=0.55$ & 42.5\% & \textbf{48.99} \\
    $\tau_\text{text}=0.6$ & 29.9\% & 48.89 \\
    $\tau_\text{text}=0.65$ & \textbf{20.1\%} & 48.62 \\
    \bottomrule
    \end{tabular}
    \caption{\textbf{Ablation on the choice of text cosine similarity threshold $\tau_\text{text}$ in CiT~\cite{xu2023cit}.} $\tau_\text{text}=0.55$, which is the default value in the original CiT paper and the value we use for our main experiments, shows the best performance. The results are from WebVid2M.}
    \label{tab:cit_threshold_ablation}
    \vskip -0.1in
\end{table}

For another ablation, we experiment with CiT without the parameter update, which we call train-free CiT, and the results can be found in Tab.~\ref{tab:cit_param_update_ablation}.
While the train-free CiT improves the computational efficiency over CiT, as it does not require any parameter update of the filtering model, its performance worsens in terms of both the average performance and the number of data sampled.

\begin{table}[t]
    \centering
    \footnotesize
    \begin{tabular}{lcc}
    \toprule
    \textbf{Filtering method} & \textbf{Clip ratio (\%)} & \textbf{Avg. perf.} \\
    \midrule
    CiT~\cite{xu2023cit} & \textbf{42.5\%} & \textbf{48.99} \\
    Train-free CiT & 51.5\% & 48.63 \\
    \bottomrule
    \end{tabular}
    \caption{\textbf{Ablation of CiT~\cite{xu2023cit} with and without filtering model parameter update.} While the Train-free CiT, which does not require the parameter update of the filtering model, improves the computation efficiency, it results in worse performance while sampling more data. The results are from WebVid2M.}
    \label{tab:cit_param_update_ablation}
    \vskip -0.1in
\end{table}

While we adapt CoLoR-Filter to operate in sample-wise manner to better suit the online video-text filtering setting, we also show the result of applying CoLoR-Filter in a batch-wise manner.
In the batch-wise CoLoR-Filter, the incoming data is first stored in a delayed buffer (batch) if it passes the video-text cosine similarity thresholding.
When the buffer is full, top $p$\% of the data within the delayed buffer is selected based on the cosine similarity difference between the finetuned and the prior models.
Tab.~\ref{tab:colorfilter_batch_version} shows the result of using batch-wise CoLoR-Filter in online video-text filtering with the buffer size of 100.
Note that while the performance slightly improves, using the batch-wise CoLoR-Filter increases the storage cost and decreases the responsiveness as it needs to wait until the buffer is full, making it less efficient and applicable in the online filtering setting.
It also introduces another hyperparameter $p$, the sampling ratio within the buffer.

\begin{table}[t]
    \centering
    \footnotesize
    \begin{tabular}{lcc}
    \toprule
    \textbf{Sampling ratio $p$} & \textbf{Clip ratio (\%)} & \textbf{Avg. perf.} \\
    \midrule
    50\% & 38.7\% & 48.46 \\
    45\% & 34.8\% & 48.46 \\
    40\% & 31.0\% & 48.58 \\
    35\% & 27.1\% & 48.67 \\
    30\% & 23.2\% & 48.67 \\
    25\% & 19.4\% & 48.44 \\
    \midrule
    Sample-wise & 56.4\% & 48.46 \\
    \bottomrule
    \end{tabular}
    \caption{\textbf{Results of using batch-wise version of CoLoR-Filter.} While using the batch-wise version slightly improves the performance, note that it incurs additional storage cost and reduces the responsiveness. The results are from WebVid2M.}
    \label{tab:colorfilter_batch_version}
    \vskip -0.1in
\end{table}

\subsection{Qualitative Analysis}
\label{subsec:qualitative_analysis}

\begin{figure*}[t]
    \begin{center}
    \includegraphics[width=\textwidth]{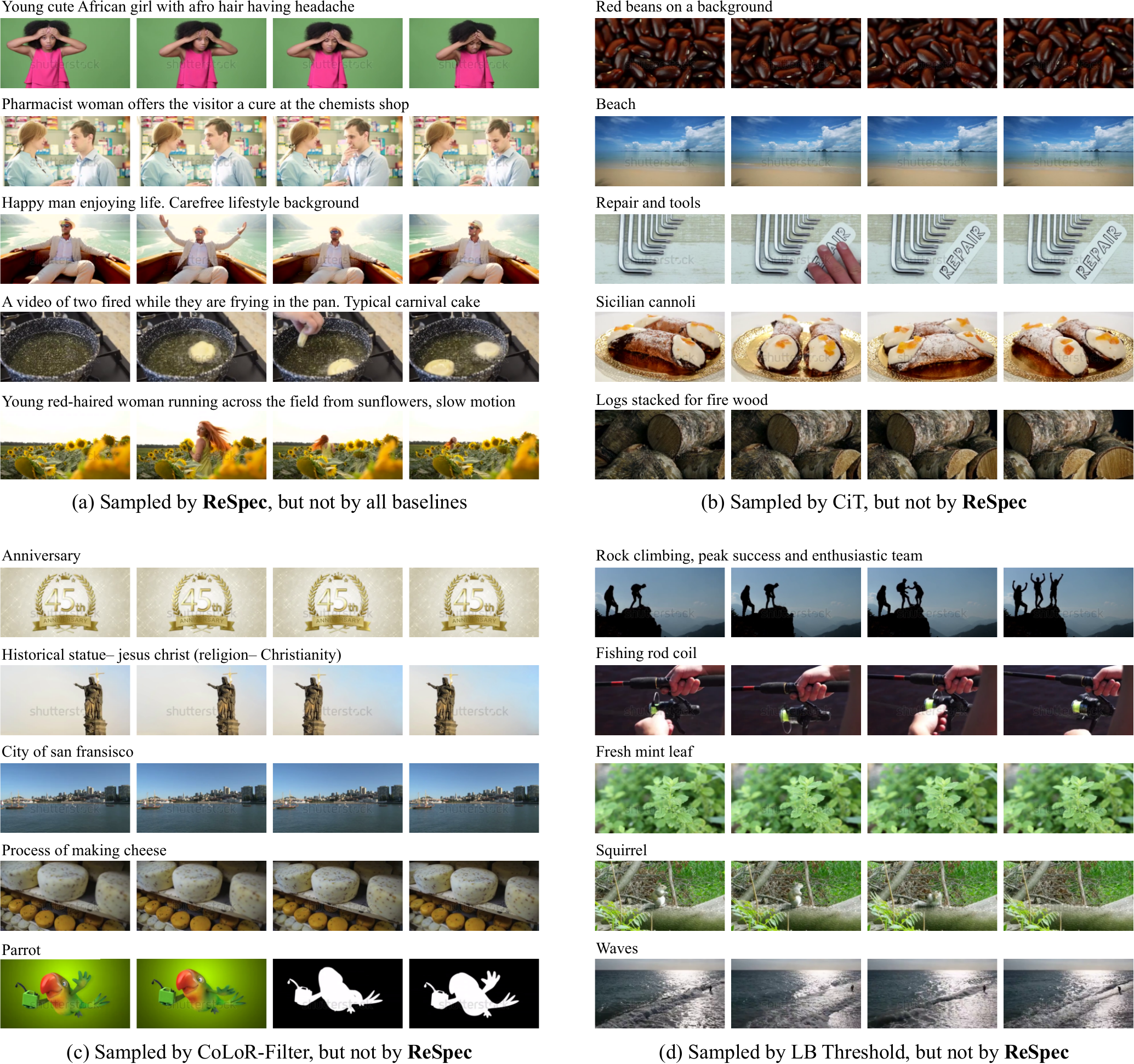}
    \caption{\textbf{Qualitative analysis}
    (a) represents samples only selected by ReSpec (ours) and not by other baselines (CiT, CoLoR-Filter). 
    (b), (c), and (d) visualize samples selected by other baselines but not selected by ours. 
    In each case, the samples selected generally ensure a certain level of alignment, relevance, and specificity. 
    While downstream relevance is shown for (b) and (c), there are noticeable shortcomings in terms of text specificity. 
    In the case of (d), there are shortcomings in both downstream relevance and text specificity.
    More samples are shown in 
    Fig.~\ref{fig:qualitative_analysis_ours}, 
    Fig.~\ref{fig:qualitative_analysis_cit}, 
    Fig.~\ref{fig:qualitative_analysis_colorfilter}
    and Fig.~\ref{fig:qualitative_analysis_lbthr}.
    } 
    \label{fig:qualitative_analysis}
    \end{center}
    \vskip -0.2in
\end{figure*}

\begin{figure*}[t] \begin{center}
    \includegraphics[width=\textwidth]{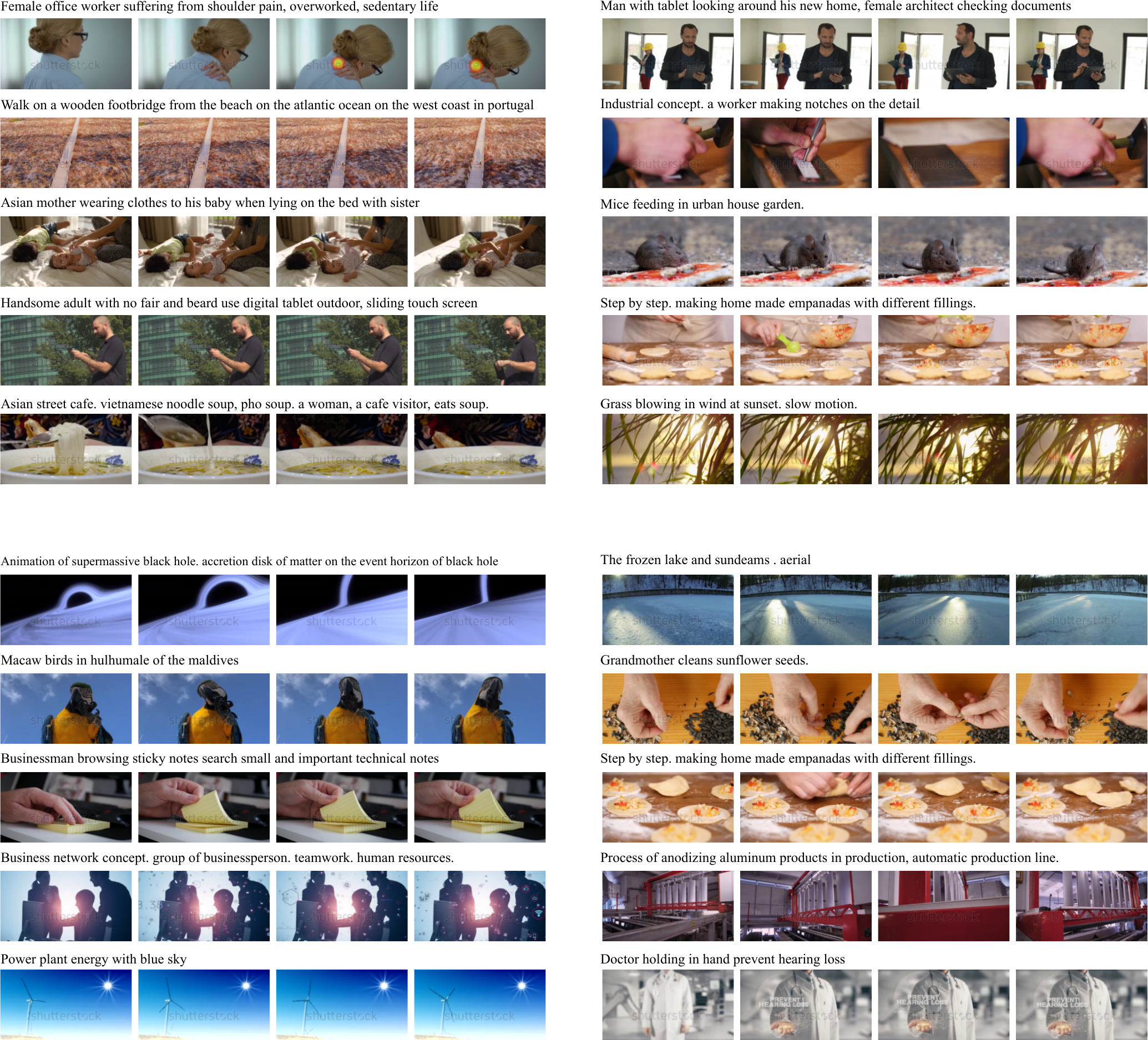}
    \caption{\textbf{Additional qualitative analysis of ReSpec}} 
    \label{fig:qualitative_analysis_ours}
\end{center} \vskip -0.2in \end{figure*}

\begin{figure*}[t] \begin{center}
    \includegraphics[width=\textwidth]{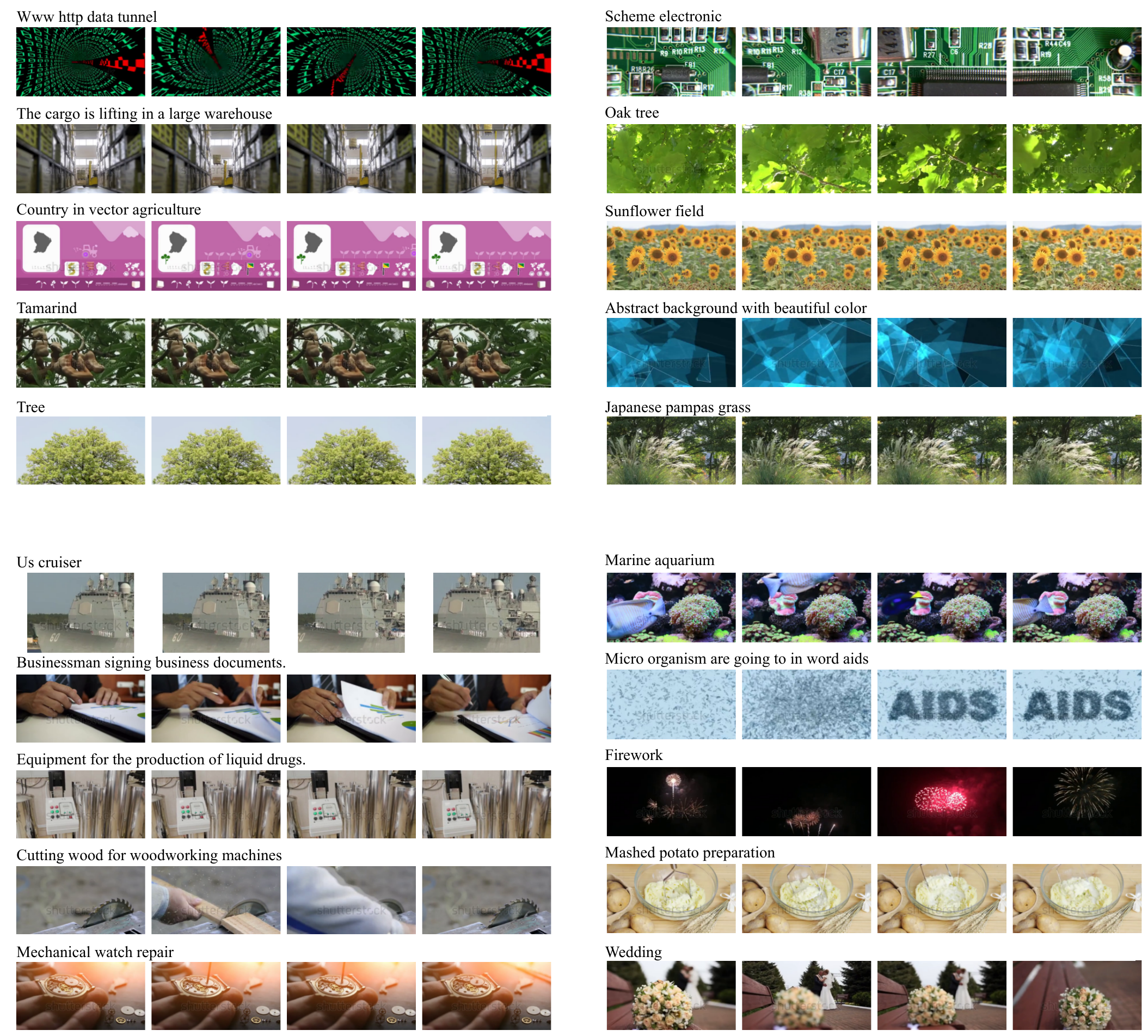}
    \caption{\textbf{Additional qualitative analysis of CiT}} 
    \label{fig:qualitative_analysis_cit}
\end{center} \vskip -0.2in \end{figure*}

\begin{figure*}[t] \begin{center}
    \includegraphics[width=\textwidth]{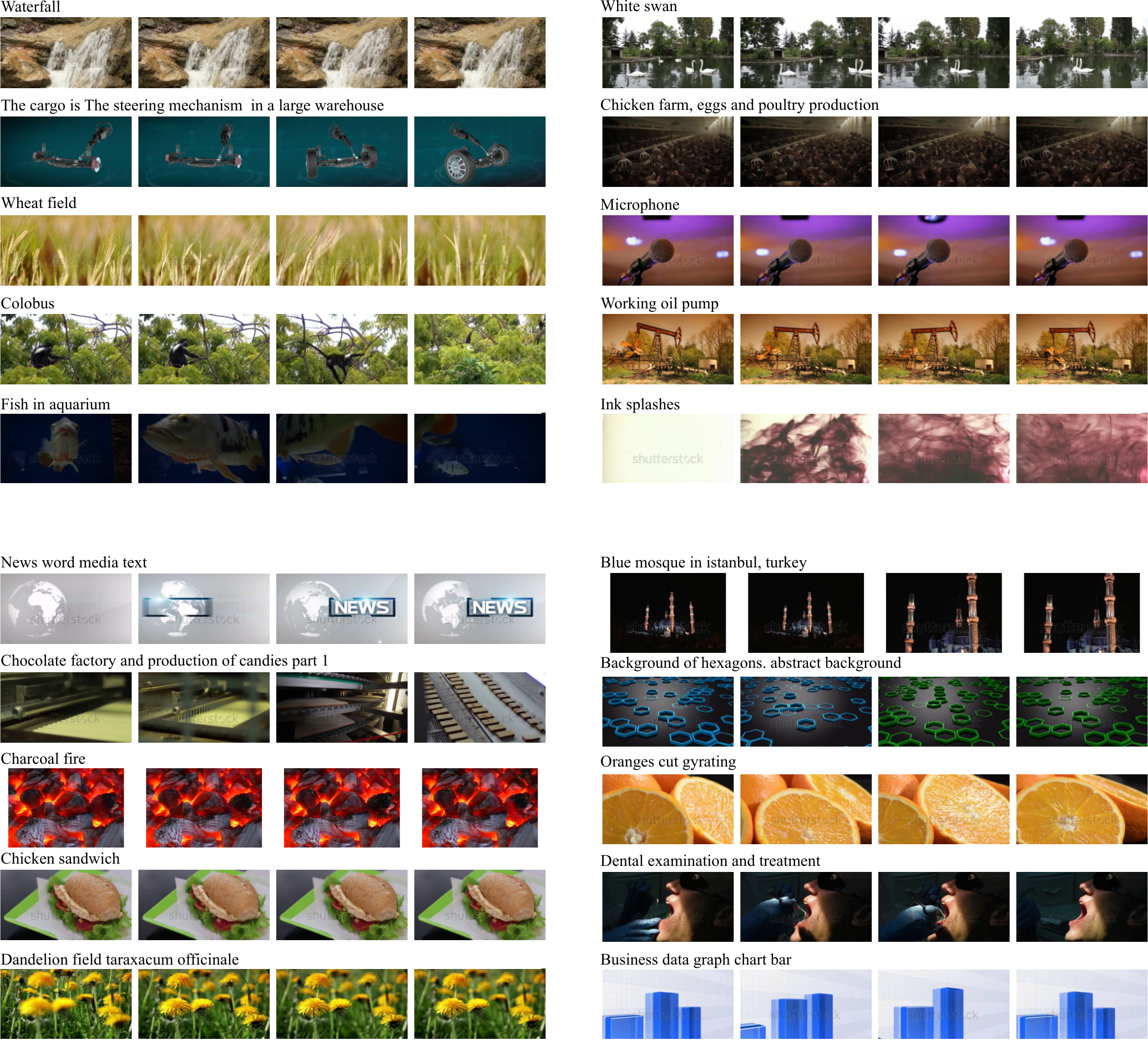}
    \caption{\textbf{Additional qualitative analysis of CoLoR-Filter}} 
    \label{fig:qualitative_analysis_colorfilter}
\end{center} \vskip -0.2in \end{figure*}

\begin{figure*}[t] \begin{center}
    \includegraphics[width=\textwidth]{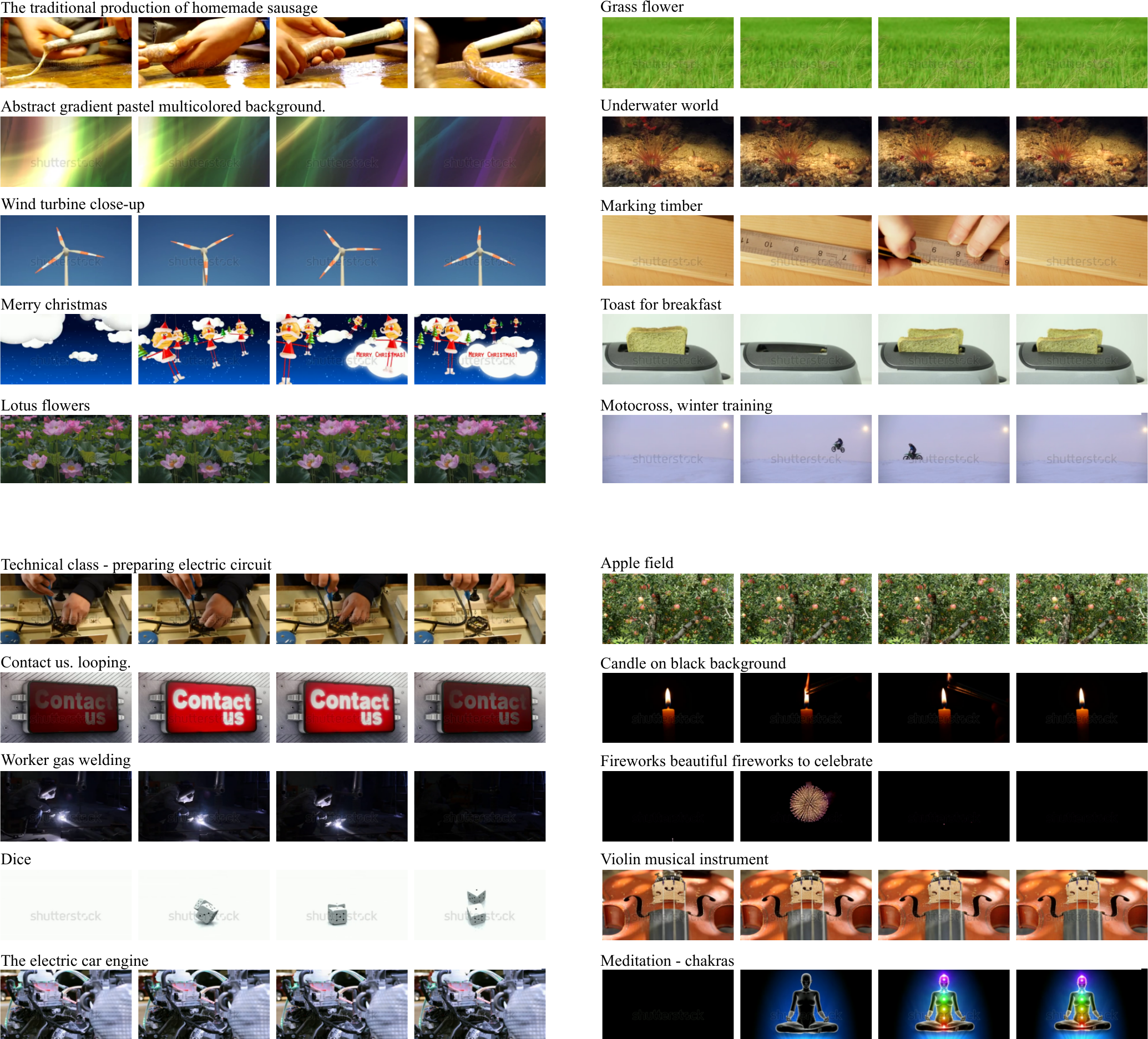}
    \caption{\textbf{Additional qualitative analysis of LB Threshold}} 
    \label{fig:qualitative_analysis_lbthr}
\end{center} \vskip -0.2in \end{figure*}

We present qualitative analyses in  
Fig.~\ref{fig:qualitative_analysis}, 
Fig.~\ref{fig:qualitative_analysis_ours}, 
Fig.~\ref{fig:qualitative_analysis_cit}, 
Fig.~\ref{fig:qualitative_analysis_colorfilter}
and Fig.~\ref{fig:qualitative_analysis_lbthr}.
Fig.~\ref{fig:qualitative_analysis}-(a) shows samples selected by ReSpec but not by any of the baselines. 
These samples are generally meaningful regarding alignment, downstream task relevance, and text specificity. 
However, the baseline fails to select these samples, indicating that the efficiency of task-aware online training is reduced.
Fig.~\ref{fig:qualitative_analysis}-(b) and (c) display samples selected by CiT and CoLoR-Filter but not by ReSpec. 
This suggests that while CiT and CoLoR-Filter maintain a certain level of downstream task relevance, 
they often select less informative samples with insufficient text specificity.
Fig.~\ref{fig:qualitative_analysis}-(d) demonstrates that the 
LB Threshold method selects data that maintains alignment but falls significantly short in relevance and text specificity.

\clearpage

\clearpage

\clearpage

\clearpage

\clearpage